\documentclass[journal]{IEEEtran}

\usepackage{tabularx,booktabs}
\newcolumntype{C}{>{\centering\arraybackslash}X} 
\usepackage{lipsum}
\usepackage{multirow}
\usepackage{booktabs}
\usepackage{float}
\usepackage{float}
\usepackage{copyrightbox}
\usepackage{subcaption}
\usepackage{graphicx}
\usepackage{todonotes}

\usepackage[]{hyperref}
\ifCLASSINFOpdf

\usepackage{amsmath}
\usepackage{amssymb}
\usepackage{graphics}
\usepackage{graphicx}
\usepackage[linesnumbered, longend, algoruled]{algorithm2e}

\SetKwInOut{Params}{Params}  
\SetKwInOut{Constants}{Constants}
\SetKwInOut{Data}{Data}
\SetKwFunction{Routing}{Dynamic Opinion Routing}
\SetKwFunction{FM}{FitMargins}
\SetKwFunction{BF}{BestFit}
\SetKwFunction{PT}{ProbabilityTransform}
\SetKwFunction{KDE}{KernelDensityEstimation}
\SetKwFunction{KDEPDF}{KDEpdf}
\SetKwFunction{CF}{Fit}
\SetKwFunction{CP}{Pdf}
\SetKwFunction{DE}{Density}
\SetKwFunction{Pr}{Prior}
\SetKwProg{Fn}{Function}{:}{end}
\SetKwComment{Comment}{$\triangleright$\ }{}

\begin{document}
	
	\title{EDC3: \textbf{E}nsemble of \textbf{D}eep-\textbf{C}lassifiers using \textbf{C}lass-specific \textbf{C}opula functions to Improve Semantic Image Segmentation}

	\author{Somenath Kuiry,
		Nibaran Das,~\IEEEmembership{Member,~IEEE,}
		Alaka Das,
		and~Mita~Nasipuri,~\IEEEmembership{Senior Member,~IEEE}
		\thanks{S. Kuiry is a research scholar of Department of Mathematics, Jadavpur University, 
			Kolkata, India             
			e-mail: skuiry.math.rs@jadavpuruniversity.in}
		\thanks{Dr. N. Das is an Associate Professor of Department of CSE, and Ms. M. Nasipuri is a Professor of Department of CSE, Jadavpur University, Kolkata 
			e-mail: nibaran.das@jadavpuruniversity.in}
		\thanks{Dr. A. Das is a Professor of Department of Mathematics, Jadavpur University, Kolkata
			e-mail: alakadas2012@gmail.com>} 
		\thanks{Dr. M. Nasipuri is a  Professor of Department of CSE, Professor of Department of CSE, Jadavpur University, Kolkata 
			e-mail: mitanasipuri@yahoo.com }
		\thanks{Manuscript received April 19, 2005; revised August 26, 2015.}}

	\markboth{}%
	{Shell \MakeLowercase{\textit{et al.}}: Bare Demo of IEEEtran.cls for IEEE Journals}

	\maketitle

	\begin{abstract}
		In the literature, many fusion techniques are registered for the segmentation of images, but they primarily focus on observed output or belief score or probability score of the output classes. In the present work, we have utilized inter source statistical dependency among different classifiers for ensembling of different deep learning techniques for semantic segmentation of images. For this purpose, in the present work, a class-wise Copula-based ensembling method is newly proposed for solving the multi-class segmentation problem. Experimentally, it is observed that the performance has improved more for semantic image segmentation using the proposed class-specific Copula function than the traditionally used single Copula function for the problem. The performance is also compared with three state-of-the-art ensembling methods.
	\end{abstract}
	

	\IEEEpeerreviewmaketitle

	\section{Introduction} \label{introduction}

\begin{figure*}[h] 
		\centering\includegraphics[width=0.4\linewidth]{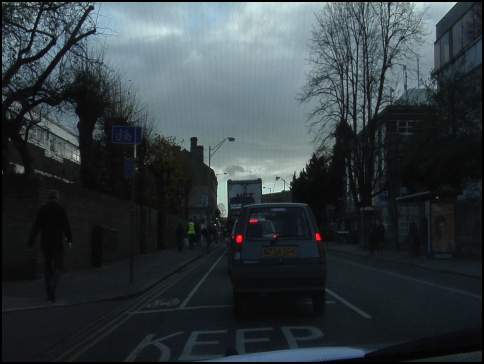} \qquad \includegraphics[width=0.4\linewidth]{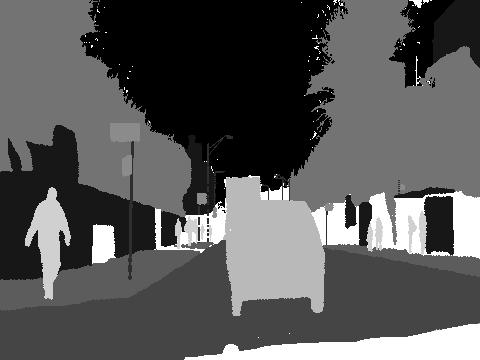}\quad
		\caption{An example of sample image along with it's segmented ground truth}
		\label{seg_image}
	\end{figure*}

	\IEEEPARstart{I}{mage} 
	segmentation techniques partition an image into multiple meaningful regions based on inherent similarities. Segmentation helps to locate objects and boundaries in an image by labeling each pixel so that pixels with the same characteristics have the same label (see Fig. \ref{seg_image}).    
	Among different types of segmentations\cite{ghosh2019understanding}\cite{shi2000normalized}\cite{achanta2010slic},  semantic segmentation \cite{long2015fully}\cite{chen2014semantic} where each pixel is associated with a class is more challenging. But semantic segmentation is very much essential in different domains like autonomous driving \cite{auto}, medical image analysis \cite{doi:10.1146/annurev.bioeng.2.1.315}\cite{forouzanfar2010parameter}, object detection and recognition \cite{delmerico2011building}, traffic control \cite{liu2018joint}, video surveillance \cite{wang2018segment} etc. Improving the accuracy of segmentation methods is always a challenge to the research community, and it has been explored heavily during the past few decades. Ensembling of multiple segmentation models is one of the popular techniques which is generally exercised by the researchers to improve the results. Various techniques are found in the literature for fusing data obtained from multiple classifiers \cite{41}. For example, majority voting \cite{42} \cite{43}\cite{44}\cite{45} is the most popular fusion technique. Other ensembling techniques like maximum, minimum, median, weighted \cite{46}, average \cite{47}, and product \cite{48}, etc. of the different classification scores are also popularly used in the literature. Rank-based method \cite{49}, the Bayesian approach \cite{44}\cite{45}, the Dempster-Shafer theory \cite{45}, fuzzy-based approaches \cite{50}\cite{51}\cite{52}, probability-based schemes like Linear Opinion Pool \cite{lop}, Beta-transformed linear opinion pool \cite{blp}, fusion based on simple logit model \cite{logit} etc. are some commonly used techniques in recent times. 
	However, none of them addresses the inter source-statistical dependence, which plays an essential role during the fusion of data, as mentioned in \cite{Ozdemir2017CopulaDependence}.   A Copula based modeling technique is proposed in the present work to use these dependencies. "Copula," which is a Latin word meaning ''a link or a bond," is a useful statistical tool for determining dependence between multivariate random variables. Copulas are functions that represent the relationship between Multivariate random variables and their marginals. This approach is more flexible than using a single multivariate distribution for multivariate data. In our proposed model, we have obtained the joint probability density of belief scores obtained from different classifiers for each pixel of an image and estimate the fused probability score using Bayes' theorem.
	
\begin{figure}[h]
		
		\centering\qquad\qquad\includegraphics[width=1\linewidth]{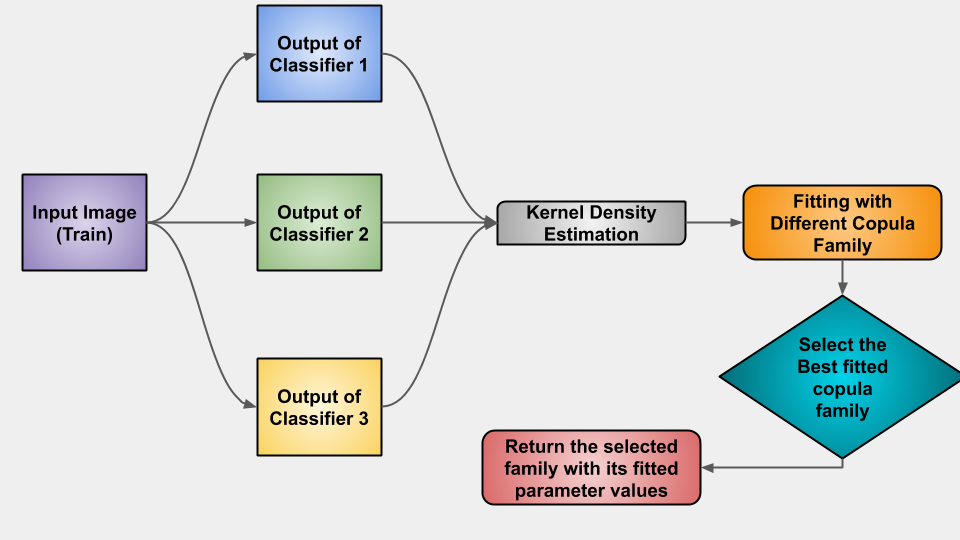} 
		\caption{Selecting the best fitted copula function using Training data}
		\label{flowchart_1}
	\end{figure}

\begin{figure}[h]
		
		\centering\qquad\qquad\includegraphics[width=1\linewidth]{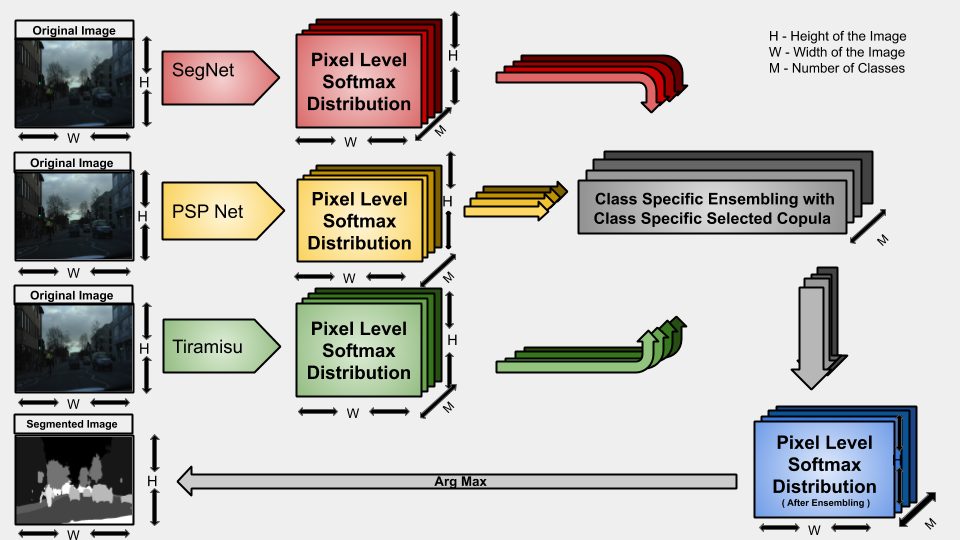} 
		\caption{ Combining the output data obtained form SegNet, PSPNet and Tiramisu with class specific copula funciton}
		\label{flowchart_2}
	\end{figure}

	Copula modelings are used in various field of research domains like hydrology \cite{laux2009modelling}, medical diagnosis \cite{Eban2013DynamicSeries}\cite{10.1371/journal.pcbi.1000577} \cite{PollanenCurrentDiagnostics}, climate and weather research \cite{Kao2009MotivatingClimate}, data mining \cite{wu2014construction} etc. \textbf{\cite{CuvelierClaytonDecomposition} \cite{Diday2005MixtureFramework}}. Though Copula is heavily used for identifying risk factors in the finance sector, its applicability is increasing day by day in other kinds of tasks such as classification  \cite{Salinas-GutierrezUsingClassification} and evolutionary computation \cite{Conant-Pablos2009PipeliningTimetabling}. Even classifier fusion approaches using Copula are also present in the literature, but to the best of our knowledge, they are mainly used for binary classification problems. \\
	Our main contributions in this paper are (1) Use of  Copula functions for the first time for ensembling of decisions of different deep learning models for semantic image segmentation, i.e., multi-class classification at pixel level; (2) Development of flexible class-specific Copula based ensemble model for a multi-class problem. A graphical overview of our proposed model is shown in Fig. \ref{flowchart_1} and Fig. \ref{flowchart_2}.\\
	The remainder of this paper is arranged as follows.  In section \ref{Problem formulation}, we formulate our problem. In section \ref{Mathematical Background}, we briefly describe all mathematical prerequisites needed for our model. In section \ref{Fitting a Copula to a Data}, we discuss various methods for fitting a copula to a data. We have presented our experimental setup, results obtained, and analysis of results has been given in section \ref{Experiments }, and the conclusion about the paper is given in section \ref{conclusion}.

	
	\section{Problem formulation} \label{Problem formulation}
	
	Image segmentation can  be treated as a pixel level classification problem.  Let, the height and width of an image  $I_{x,y}$ is represented by $H$ and $W$.  After segmentation,  a model or a classifier  gives an output of size $H\times W \times M$, where $M$ is the total number of associated classes for each pixel. So for each pixel, the model returns an array of  $M$ values, which represents the model's belief scores or confidence scores about that pixel to be in those $M$ classes.\\
	Suppose, there are  $L$ number of segmentation models or classifiers. So for each pixel of an image, these models generate  belief scores $p_{i}^{(j)},i=1,2\cdots,L;j=1,2,\cdots M$, which represents their beliefs on a single class $j$. Our objective is to find a function $g :[\,0,1]\,^{L}\longrightarrow[\,0,1]\,$ to compute the fused belief score $p^{(j)} := g(\,p^{(j)}_{1},p^{(j)}_{2},.....,p^{(j)}_{L})\,$, which will be the final belief score for the class $j$.
	To evaluate the mapping $g$, we use $T$ number of training examples and their ground truth labels for each pixel.
	Most of the fusion techniques assume that participating classifiers are independent of each other. But in real scenarios, there might have  some statistical dependence among those classifiers when they are trained on same dataset \cite{Ozdemir2017CopulaDependence}. \\
	The fundamental concept of our strategy is  to treat each confidence score $p_{i}^{j}$ $i = 1,2\cdots L;j=1,2,\cdots M$ as an observation and to construct the joint likelihood of belief scores under class $j$, $f\left(p^{(j)}_{1},p^{(j)}_{2},.....,p^{(j)}_{L} \mid \text{Class }  j\right)$ for each pixel using training samples. After that, Bayes' theorem is used to obtain the fused belief score $p^{j}$ as
	\begin{align*}
	p^{j}  &:= g (p^{(j)}_{1},p^{(j)}_{2},.....,p^{(j)}_{L}) \\
	&:= P\left(\text{Class } j \mid p^{(j)}_{1},p^{(j)}_{2},.....,p^{(j)}_{L}\right) \nonumber \\ 
	& \propto f\left(p^{(j)}_{1},p^{(j)}_{2},.....,p^{(j)}_{L}\mid \text{Class }  j\right) \times P\left(\text{Class } j\right) \label{eq1}
	\end{align*}
	The challenge is to evaluate the joint likelihood $f\left(p^{(j)}_{1},p^{(j)}_{2},.....,p^{(j)}_{L}\mid \text{Class } j\right)$ under unknown statistical dependence. To solve that,  Copula function, a popular statistical tool,  which is capable of modeling these kinds of dependencies is used here. 
	In the next section, the mathematical background of Copula is revisited.
	
	\section{Mathematical Background} \label{Mathematical Background}
	
	\subsection{Definition :} \label{Definition }
	Let $X_{1},X_{2},....,X_{N}$ be $N$ random variables such that their marginal distributions are continuous. We assume that $F_{i}(x_i)$ is the marginal distribution of $X_{i}$ i.e. $F_{i}(x_i) = P[\,X_{i}\leq x_i ]\,$ for $i=1,2 \dots N$. The Copula of $(X_{1},X_{2},...,X_{n})$ is defined as the joint cumulative distribution over $(U_{1},U_{2},...,U_{n})$ as given below :
	\begin{align}
	C\left(u_{1},u_{2},...,u_{n}\right) = P\left(U_{1}\leq u_{1},U_{2}\leq u_{2},...,U_{n}\leq u_{n}\right) \nonumber
	\end{align}
	where $U_{i} = F_{i}(x_{i})$.
	
	\subsection{Sklar's Theorem :} \label{Sklar's Theorem}
	Let $F(x_{1},x_{2},....,x_{N})$ be a $N$ dimensional cumulative distribution function over real valued random variable with margins $F_{1},F_{2},...,F_{N}$. Then there exists a Copula $C : [0,1]^{N} \mapsto [\,0,1]\, $ such that for all $x_{1},x_{2},...,x_{N} \in [-\infty,\infty]$
	$$F(x_{1},x_{2},....,x_{N}) = C(F_{1}(x_{1}),F_{2}(x_{2}),,...F_{N}(x_{N}))  $$
	
	Furthermore if the margins $F_{i}(x_{i})$ are continuous then $C$ is unique. Otherwise we can determine $C$ uniquely on $Ran(\,F_{1})\,\times Ran(\,F_{2})\,\times.....\times Ran(\,F_{N})\,$, where $Ran(\,F_{i})\,$ denotes the Range of $F_{i}$.
	
	\subsection{Copula Density :} \label{Copula Density}
	Copula function is a multivariate distribution function. Thus we can obtain its density function by partially differentiating with respect to all its variables as :
	$$
	c(\,u_{1},u_{2},...,u_{N})\, = \frac{\partial ^{N}C(\,u_{1},u_{2},...,u_{N})\,}{\partial u_{1}\partial u_{2}\cdots \partial u_{N}}
	$$ 
	Here the copula function $C$ is assumed to be sufficiently differentiable with respect to all its arguments.\\
	To established the relation between copula density function and multivariate density function, we differentiate the above equation as follows:
	\\
	\begin{align}
	f(\,x_{1},x_{2},...,x_{N})\, & = \frac{\partial ^{N}F(\,x_{1},x_{2},...,x_{N})\,}{\partial x_{1}\partial x_{2}\cdots \partial x_{N}} \nonumber \\
	& = \frac{\partial ^{N}C(\,F_{1}(\,x_{1})\,,F_{2}(\,x_{2})\cdots F_{N}(\,x_{N})\,)\,}{\partial x_{1}\partial x_{2} \cdots \partial x_{N}}\nonumber \\
	&= c\left(F_{1}(\,x_{1})\,,F_{2}(\,x_{2})\cdots F_{N}(\,x_{N}) \right) \prod_{i=1}^{N} f_{i}(\,x_{i})\, 
	\end{align}
	where $f_{i}$ being marginal density of $X_{i}$ for all $i$.
	This copula density is significant for fitting data to a copula and also to determine their likelihood. Some of the important copula functions are discussed below.
	
	\subsection{Copula Families :} \label{Copula Families}
	There are different kinds of copula classes (collection of copula families which have similar properties) present in the literature, viz, fundamental, elliptical, archimedean, and hierarchical archimedean. A brief introduction of different copula families is given in this section.
	\begin{itemize}
		\item Fundamental Copulas :
		There are three fundamental copula families namely comonotonicity (represents perfect positive dependence), countermonotonicity (represents perfect negative dependence) and independent (represents independence) copulas The simplest is the Independent copula (see Fig. \ref{independent Copula} ), which has the following form
		\begin{equation}
		C(u_{1},u_{2}\cdots u_{N}) = \prod_{i=1}^{N} u_{i} \nonumber
		\end{equation}
		\begin{figure}[h]
			\centering\qquad\includegraphics[width=0.4\linewidth]{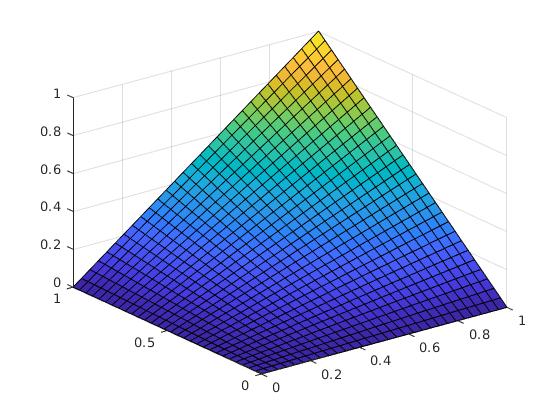} \quad \includegraphics[width=0.4\linewidth]{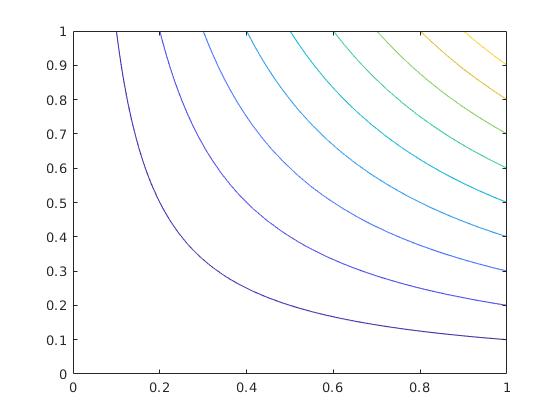}
			\caption{CDF and countour plot of Independent Copula}
			\label{independent Copula}
		\end{figure}
		\item Elliptical Copulas :
		Copulas which describes the dependencies of elliptical multivariate distribution are called Elliptical Copulas. The copula families belonging to this class are gaussian copula and student-t copula.
		\begin{itemize}
			\item Gaussian copula :
			describes the multivariate normal (Gaussian) distribution and has the following form:
			\begin{equation}
			\resizebox{.4\textwidth}{!} 
			{
				$ C(u_{1},u_{2} \cdots u_{N}) = \Phi_{\Sigma}(\,\Phi^{-1}(u_{1}),\Phi^{-1}(u_{2})\cdots \Phi^{-1}(u_{N}))\,\nonumber $
			}
			\end{equation}
			Where $\Phi_{\Sigma}$ is cumulative distribution function of multivariate normal distribution with correlation matrix $\Sigma$ and $\Phi^{-1}$ is the quantile function of normal distribution. Gaussian copula can describe a variety of dependences depending on its parameter. 
			For example in bivariate case $\Sigma$ is is $2\times 2$ matrix with a single parameter $\rho = \Sigma_{1,2} = \Sigma_{2,1}$. For $\rho = -1$ it represents countermonotonicity copula, for $\rho = 1$ it represents Comonotonicity copula, for $\rho = 0$ it represents independent copula etc.
			\begin{figure}[h]
				\centering\qquad\includegraphics[width=0.4\linewidth]{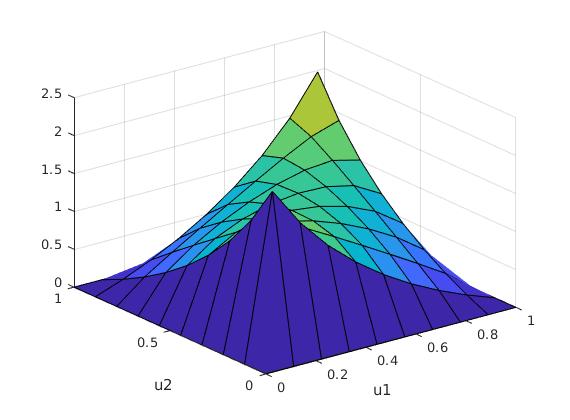} \quad \includegraphics[width=0.4\linewidth]{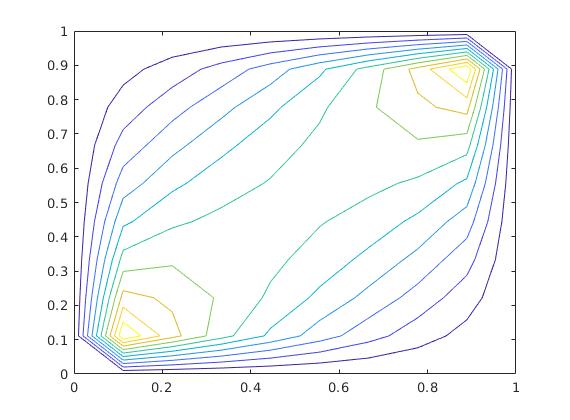} \\
				\centering\qquad\includegraphics[width=0.4\linewidth]{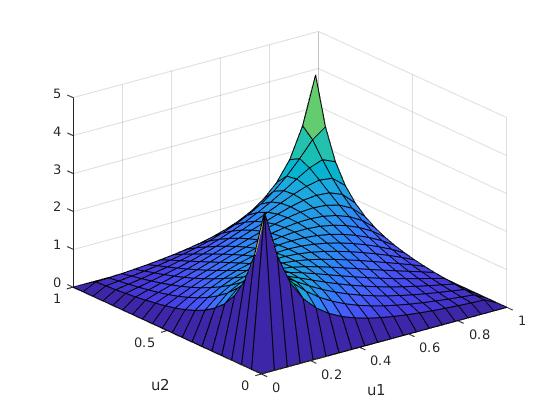} \quad \includegraphics[width=0.4\linewidth]{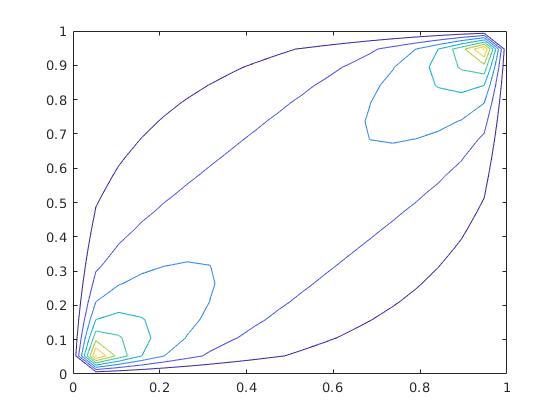}
				\caption{CDF and countour plots of Gaussian Copula and Student-t Copula respectively}
				\label{gaussian Copula}
			\end{figure}
			\item Student-t Copula : Similar to Gaussian Copula the Student-t copula can be defined as follows:
			\begin{equation}
			\resizebox{.4\textwidth}{!} 
			{
				$C(u_{1},u_{2} \cdots u_{N}) = t_{\nu,\Sigma}(\,t_{\nu}^{-1}(u_{1}),t_{\nu}^{-1}(u_{2}) \cdots t_{\nu}^{-1}(u_{N}))\,\nonumber$
			}
			\end{equation}
			Where $t_{\nu,\Sigma}$ is the cumulative distribution function(CDF) of multivariate Student-t distribution with correlation matrix $\Sigma$ and degrees of freedom $\nu $ and $t_{\nu}$ is the CDF of univariate Student-t distribution with degrees of freedom $\nu$. As $\nu \rightarrow \infty$ the Student-t copula converge to Gaussian copula(The difference becomes negligible after $\nu \geq 30$ \cite{gtot}). Student-t copula is mostly used in finance studies since it exhibits best fit than other families. And example of Gaussian Copula and Student-t copula is shown in Fig. \ref{gaussian Copula}. 
			
		\end{itemize}
		\item Archimedean Copula: copulas generated by archimedean generator (with one or more parameters) are called Archimedean Copula. Depending on different parameter values, different copulas can be obtained. For a single-parameter family, there exist 22 copulas \cite{differentcopula}; among those, we have chosen here three, which are most commonly referred to in literature. 
		\begin{itemize}
			\item Clayton Copula: It can describe the dependence in the lower tail (as shown in Fig. \ref{Clayton Copula}), and that is why it is used in finance to detect correlated risk. This type of copulas is represented as :
			\begin{equation}
			\resizebox{.4\textwidth}{!} 
			{
				$ C(u_{1},u_{2} \cdots u_{N}) = \bigg(\sum_{i=1}^{N} u_{i}^{-\theta} - 1\bigg)^{-\frac{1}{\theta}},\theta \in [-1,\infty)\backslash \{0\} \nonumber $
			}
			\end{equation}
			\item Frank Copula : It can describe symmetric dependence (as shown in Fig. \ref{Clayton Copula}) also unlike Clayton it can describe positive as well as negative dependence. It has the following form :
			\begin{align}
			C&(u_{1},u_{2} \cdots u_{N}) \hspace{4 cm} \nonumber \\ 
			&  = -\theta^{-1}\log \left[1+(e^{-\theta}-1)^{-1}\prod_{i=1}^{N} (e^{-\theta u_{i}}-1)\right] \nonumber \\
			& \qquad \qquad \text{Where } \theta \in \mathbb{R} \backslash \{0\}\nonumber
			\end{align}    
			
			\begin{figure}[h]
				\centering\qquad\includegraphics[width=0.4\linewidth]{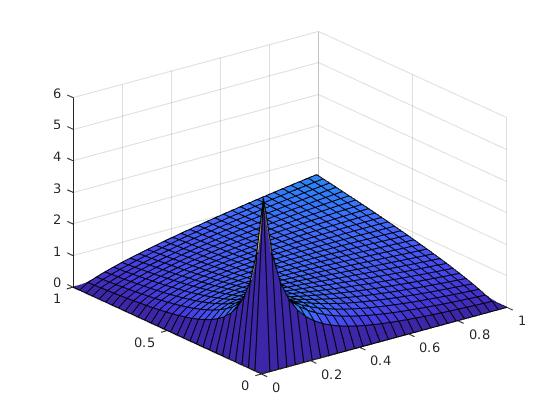} \quad \includegraphics[width=0.4\linewidth]{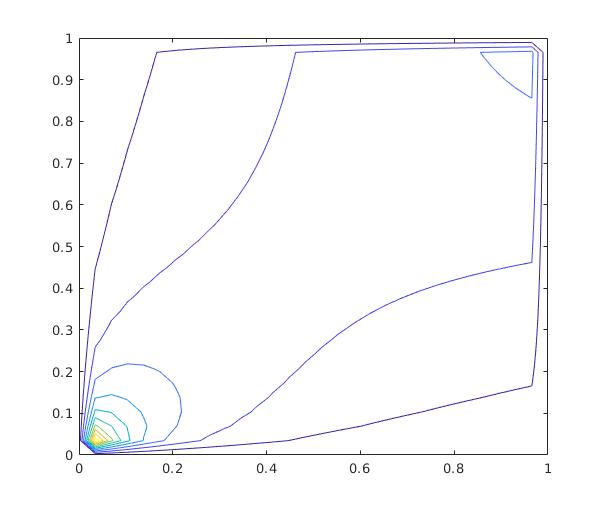} \\
				\centering\qquad\includegraphics[width=0.4\linewidth]{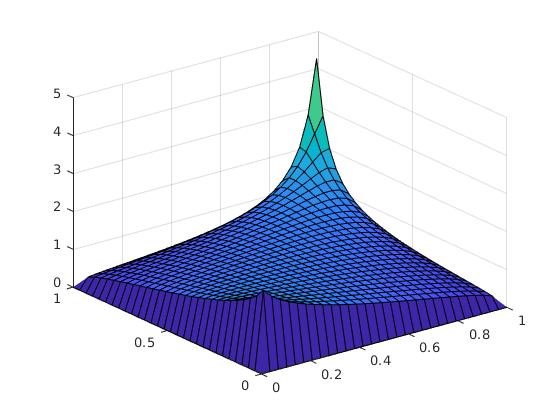} \quad \includegraphics[width=0.4\linewidth]{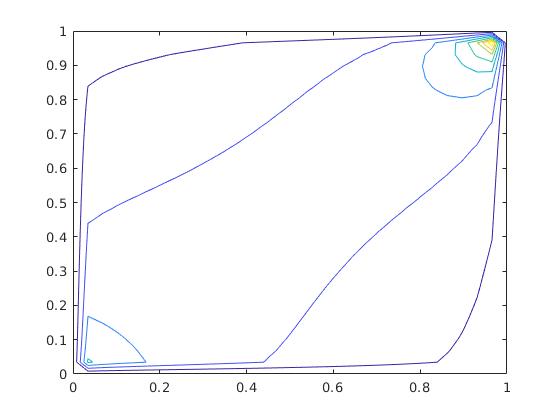} \\
				\centering\qquad\includegraphics[width=0.4\linewidth]{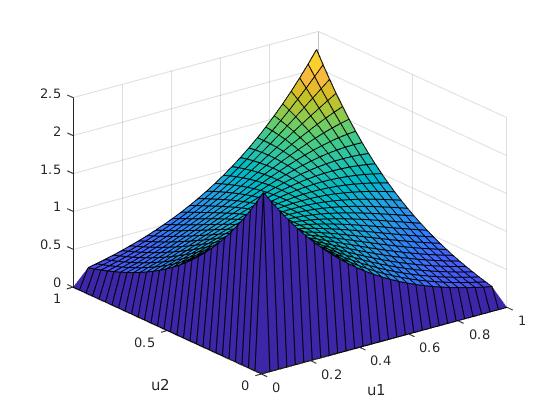} \quad \includegraphics[width=0.4\linewidth]{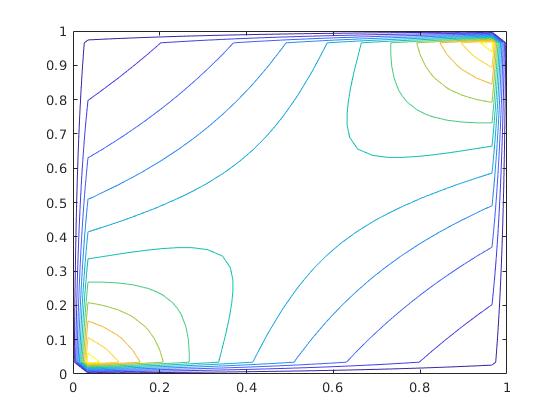}
				\caption{CDF and countour plots of clayton, frank and gumbel Copula respectively}
				\label{Clayton Copula}
			\end{figure}
			
			\item Gumbel Copula : It can describe asymmetric dependence (as in Fig. \ref{Clayton Copula}). Like clayton it cannot represent negative dependence. It has the following form:
			
			\begin{equation}
			\resizebox{.4\textwidth}{!} 
			{
				$ C(u_{1},u_{2} \cdots u_{N})  = exp\Bigg(-\bigg[\sum_{i=1}^{N}(-\log u_{i})^{\theta}\bigg]^{\frac{1}{\theta}}\Bigg), \theta \in [1,\infty)\nonumber $ 
			}
			\end{equation}

		\end{itemize}
		\item Hierarchical Archimedean Copula: Archimedean Copula is used in various cases, but they have some limitations like exchangeability of variables, less number of parameters. So, they are not quite useful for higher dimensions. In higher dimensions, a generalized version of Archimedean copulas is usefully called Hierarchical Archimedean Copula(HAC). HACs are constructed by composing two or more simple Archimedean copulas, and they can describe a wide range of dependence structures. The details can be found elsewhere \cite{hac}\cite{hac2}. 
	\end{itemize}
	
	\section{Fitting a Copula to a Data } \label{Fitting a Copula to a Data}
	To fit a Copula to a $d$ -dimensional data, we need to estimate the parameters of that Copula, which will describe the multivariate distribution of those samples. For example, in the case of elliptical copulas, we need to estimate the correlation matrix of those samples. Similarly, we need to estimate the degree of freedom of the samples for student-t Copula. For Archimedean Copulas, we have to estimate only the generator. In the next section, we will discuss various fitting methods \cite{ml}\cite{ml2}\cite{ifm}\cite{ifm2}.
	
	\subsection{The Maximum Likelihood(ML) method \cite{ml} :} \label{ml method}
	In this method we estimate both copula parameters and marginal parameters. Given a sample $(X_{1}^{t},X_{2}^{t} \cdots X_{N}^{t})_{t=1}^{T}$ for $T$ training examples we get the likelihood function $\mathcal{L}(c,\tau)$ from copula density function as :
	
	\begin{align}
	\mathcal{L}(c,\tau) &= \prod_{t=1}^{T} f^{t}(c,\tau) \nonumber \\
	& = \prod_{t=1}^{T}(\ c(\,F_{1}(\,x_{1})\,,F_{2}(\,x_{2})\,,...F_{N}(\,x_{N})\,)\, \prod_{i=1}^{N} f_{i}^{t}(\,x_{i})\  )\ \label{eq5}
	\end{align}
	Taking $\log$ on both side of (\ref{eq5}) we get the log-likelihood function as ,
	\begin{align}
	&\mathcal{L}(c,\tau) \nonumber \\
	&= \log(\mathcal{L}(c,\tau)) \nonumber\\
	&= \log \Bigg(\prod_{t=1}^{T} \bigg( c(F_{1}(x_{1}),F_{2}(x_{2})\cdots F_{N}(x_{N}), \tau ) \prod_{i=1}^{N} f_{i}^{t}(x_{i})  \bigg) \Bigg) \nonumber \\
	& = \sum_{t=1}^{T} \log \bigg(c(F_{1}(x_{1}),F_{2}(x_{2})\cdots F_{N}(x_{N}) , \tau ) \prod_{i=1}^{N} f_{i}^{t}(x_{i})  \bigg) \nonumber\\
	& = \sum_{t=1}^{T} \log \bigg(c\big(F_{1}(x_{1}),F_{2}(x_{2})..F_{N}(x_{N} ), \tau \big) \bigg) \nonumber\\ & \hspace{4 cm} + \bigg(\sum_{t=1}^{T} \sum_{i=1}^{N} \log f_{i}^{t}(x_{i}) \bigg) \label{eq6} 
	\end{align}
	Hence the ML estimator is 
	\begin{align}
	\hat{\mathcal{L}} (\hat{c},\hat{\tau}) = \arg \max \mathcal{L}(c,\tau) \label{eq7}
	\end{align}
	If marginals are known then we can rewrite (\ref{eq7}) as 
	\begin{align}
	\hat{\mathcal{L}} (\hat{c},\hat{\tau}) & = \arg \max \mathcal{L}(c,\tau)  \nonumber \\
	& = \arg \max_{\substack{\tau \in \tau_{\mathcal{C}} \\ 
			c \in \mathcal{C}}} \sum_{t=1}^{T} \log \bigg(c\big(F_{1}(\,x_{1})\,,F_{2}(\,x_{2})\,\dots F_{N}(\,x_{N} )\, , \tau \big) \bigg) 
	\end{align}
	If the marginals are not known then the ML estimator becomes 
	\begin{align}
	&\hat{\mathcal{L}} (\hat{c},\hat{\tau},\hat{f_{1}}(x_{1}),\hat{f_{2}}(x_{2}),\dots \hat{f_{N}}(x_{N})) \nonumber \\
	& = \arg \max_{\substack{\tau \in \tau_{\mathcal{C}} \\ 
			c \in \mathcal{C} \\ f_{i}(x_{i}) \in \mathcal{F}}} \Bigg[ \sum_{t=1}^{T} \log \bigg(c\big(F_{1}(\,x_{1})\,,F_{2}(\,x_{2})\,..F_{N}(\,x_{N} )\, , \tau \big) \bigg) \nonumber \\
	& \hspace{3 cm} + \bigg(\sum_{t=1}^{T} \sum_{i=1}^{N} \log f_{i}^{t}(\,x_{i})\  \bigg) \Bigg] \label{eq9}
	\end{align}
	Due to the complex structure of equation (\ref{eq9}) and presence of huge amount of data ML method is not suitable. Thus Inference Function for Margins Method(IFM) is adopted to solve such problems.
	
	\subsection{The Inference Function for Margins Method (IFM) \cite{ifm} :} \label{ifm}
	The above equation (\ref{eq9}) can be simplified by estimating the marginals first then using those marginals to estimate the copula parameters. Equation (\ref{eq9}) can be rewritten as :
	\begin{align}
	\mathcal{L}&(c,\tau,\xi) \nonumber\\ 
	& = \sum_{t=1}^{T} \log \bigg(c\big(F_{1}(x_{1},\xi_{1}),F_{2}(x_{2},\xi_{2})\dots F_{N}(x_{N},\xi_{N} ) , \tau \big) \bigg) \nonumber \\
	&  \hspace*{2 cm}  + \bigg(\sum_{t=1}^{T} \sum_{i=1}^{N} \log f_{i}^{t}(x_{i},\xi_{i}) \bigg)  \label{eq10}
	\end{align}
	Here $\xi = (\xi_{1},\xi_{2} \dots \xi_{N}) $ represents the parameters of marginals and $\tau $ represents the parameter of copula.
	Now in this method we first estimate $\xi$ from the second part of (\ref{eq10}) i.e.
	\begin{align}
	\hat{\xi_{i}} 
	& = \arg \max_{\xi_{i}} \bigg(\sum_{t=1}^{T} \log f_{i}^{t}(x_{i},\xi_{i}) \bigg) \label{e11}
	\end{align}
	After that using (\ref{e11}) we estimate the first part of equation (\ref{eq10}) as :
	\begin{align}
	\hat{\tau} & = \arg \max_{\tau} \sum_{t=1}^{T} \log \bigg(c\big(F_{1}(x_{1},\hat{\xi_{1}}),F_{2}(x_{2},\hat{\xi_{2}})\cdots\nonumber \\
	& \hspace*{2 cm} \cdots F_{N}(x_{N}, \hat{\xi_{N}} ), \tau \big) \bigg) 
	\end{align}
	Hence the IFM estimator will be $(\hat{\xi},\hat{\tau})$. This method is asymptotically equivalent to the ML method. 
	
	\subsection{Estimating Marginals :} \label{Estimating Marginals}
	In most of the cases the marginals $f_{i}(x_{i})$ are unknown and have to be estimated for IFM method. To do this we have adapted Kernel Density Estimation(KDE) to fit our data. KDE is a nonparametric technique for estimation of probability density of any random variable. Let $x_1,x_2 \cdots x_k$ be $k$ samples drawn from an unknown distribution. Then for any value $x$ the fromula for KDE is $$\widehat{f}_{h}(x)=\frac{1}{k h} \sum_{i=1}^{k} K\left(\frac{x-x_{i}}{h}\right)$$ Where $K(.)$ is the kernel smoothing function and $h$ is the bandwidth which controls the smoothness of the resulting density curve. For our data we have used Normal kernel or Gaussian kernel in KDE. The formula for gaussian kernel is given by $$K(x )=\frac{1}{\sqrt{2 \pi}} \exp \left(-\frac{x^{2}}{2}\right)$$
	
	\subsection{Measuring the fit :} \label{Measuring the fit}
	There are various Measuring statistics available in the literature. Some of them are Log-likelihood(LL), Akaike Information Criteria (AIC) \cite{Burnham2004MultimodelSelection} \cite{Aho2014ModelBIC}, Bayesian Information Criterion (BIC) \cite{Burnham2004MultimodelSelection} \cite{Aho2014ModelBIC}, Average Kolmogorov-Smirnov distance(AKS), Cramer-von-Mises distance etc. Out of these, we have used Log-likelihood, AIC, and BIC for our experiment.
	
	\section{Experiments :} \label{Experiments }
	To validate the effectiveness of the proposed class-specific Copula based ensemble method i.e EDC3( Ensemble of Deep classifiers using class-specific Copula function), we have evaluated the technique on two publicly available datasets CamVid \cite{camvid}, a road scene video sequence consisting of $601$ frames  and ICCV09 \cite{iccv09}, a collection of urban and rural scene images consisting of $715$ frames. The observed results of the ensembling method are compared with the individual classifiers along with the ensemble of those classifiers using different single copula functions. The proposed technique is also compared with three other state-of-the-art ensembling methods to prove the superiority of the proposed method. The details of the experimental results are shown in Table \ref{main_table},\ref{big_main_table},\ref{main_iccv09} and \ref{big_iccv09}. 
	
	\subsection{Data-sets :} \label{datasets}
	
	\subsubsection {CamVid}CamVid \cite{camvid} is a road scene video sequence consisting of $367$ frames of train set, $101$ frames of validation set and $233$ frames of test set. In our experiment, we set the dimension of each frame to  $360 \times 480$, which is half of the original dimension. The ground truth of Camvid has $11$ semantic segmentation classes, namely Sky, Building, Column-pole, Road, Sidewalk, Tree, Sign-symbol, Fence, Car, Pedestrian, Bicyclist, and one void class. We have trained our  base deep models, i.e., SegNet, PSPNet, and Tiramisu, with these $12$ semantic classes. 
	
	\subsubsection{ICCV09} We have also used another dataset ICCV09 \cite{iccv09}, which consists of $715$ images of urban and rural scenes assembled from a collection of public image datasets. There are $572$ images for training and $143$ images for validation purposes. It has a total of $9$ semantic segmentation classes, namely Sky, Tree, Grass, Ground, Building, Mountain, Water, Object, and one unknown class. The resolution of each image here is  $ 240 \times 340$, but in our experiment, we have resized each image to $360 \times 480$. 
	
	\subsection{Experimental Setup :} \label{experiment}
	For segmentation of images on  CamVid(12) and ICCV09 dataset, three popular deep learning-based segmentation models  namely SegNet \cite{BadrinarayananSegNet:Segmentation}, PSPNet \cite{ZhaoPyramidNetwork} and Tiramisu\cite{JegouTheSegmentation} are used. SegNet \cite{BadrinarayananSegNet:Segmentation} is an encoder-decoder architecture with forwarding pooling indices which has the potential to delineate boundaries between classes. On the other hand, PSPNet \cite{ZhaoPyramidNetwork} is a pyramid spatial pooling module based architecture for aggregating contextual information, which is popularly used for image segmentation nowadays. The one hundred layers Tiramisu \cite{JegouTheSegmentation} is a profound architecture made of fully convoluted DenseNets which helps to boost segmentation performance significantly.\\
	Each of the above models returns its belief score as a probability distribution across all the classes for each pixel of the input image using a softmax function in our case.  Hence, if the input size is $H \times W$, then the Classifier output size will be $H \times W \times M$, Where $M$ is the total number of classes. \\
	The next step is to determine which Copula will be the best fit for the data (for each class) during the ensemble procedure. To do that, we fit our data for each class to five popular elliptic and Archimedean copulas, namely 1)Gaussian, 2)Students-t, 3)Clayton, 4)Gumbel, and 5) Frank and determine the LL, AIC and BIC statistics \cite{differentcopula} \cite{Burnham2004MultimodelSelection} \cite{Aho2014ModelBIC}. After evaluating those statistics, the best-fitted Copula will be chosen for each class. At the end of this procedure, all the selected class-specific best copulas are used to estimate their parameters. The final class-specific fused distribution will be determined using these parameters and validation data (Classifier outputs on validation samples for each class as a probability distribution across total classes).  The whole approach is presented briefly in \textbf{Algorithm 1}.
	\begin{algorithm*}
		\caption{Copula Ensembling}

		\Constants{L = Number of Classifiers \\
			M = Number of segmentation classes\\
			P = Total Number of Pixel for Training Images \\
			Q = Total Number of Pixel for Validation Images\\
			\textbf{Family} = The best fitted copula function corresponding \\marginals for the given data }
		
		\Data{
			$[X_{m}^{l}]_{P \times 1}$ = Pixel-level Probability Distribution for class $m$, from classifier $l$.\\
			$[XT_{m}^{l}]_{Q \times 1}$ = Pixel-level Probability Distribution for\\ class $m$, from classifier $l$.}
		
		\KwResult{$R_{m}$ = Pixel-Level Probability Distribution after ensembling for class $m$.}
		\For{$m=1$ \KwTo $M$}{
			{
				$X = [X_{m}^{1} X_{m}^{2} \dots X_{m}^{L}]$\\
				
				U = \KDE{X}\\
				Copula-Parameters = \CF{\textit{Family},U} \Comment*[r]{by IFM method}
				$XT = [XT_{m}^{1} XT_{m}^{2} \dots XT_{m}^{L}]$\\
				UT = \KDE{XT}\\
				$A_{l}$ = \CP{\textit{Family},UT,Copula-Parameters}\\
				$R_{m} = A_{m}$ * \KDEPDF{TX} * \Pr{m} \Comment*[r]{by equation (1)}    
			}
		}

	\end{algorithm*}   
	To fit data to a copula, we have used the function \CF based on the \textbf{IFM} method (see section \ref{ifm} )in Algorithm 1, which returns an estimate of parameters of the given copula family. The term \textit{Family} is used here to denote the best-fitted Copula family for given input data, which are determined empirically using their fitting statistics(AIC, BIC, LL) after fitting with some well-known copula families. The copula family fitting statistics for each class of training data on the CamVid dataset are presented in the Appendix. The function \KDE is the non-parametric marginal estimation of our given data. 
	The \CP function returns the probability density function of the selected copula family for the estimated copula parameters at the test data samples. 
	The function \KDEPDF calculates the marginal density of the given samples.
	\begin{figure*}[h]
		\centering\includegraphics[width=1\linewidth]{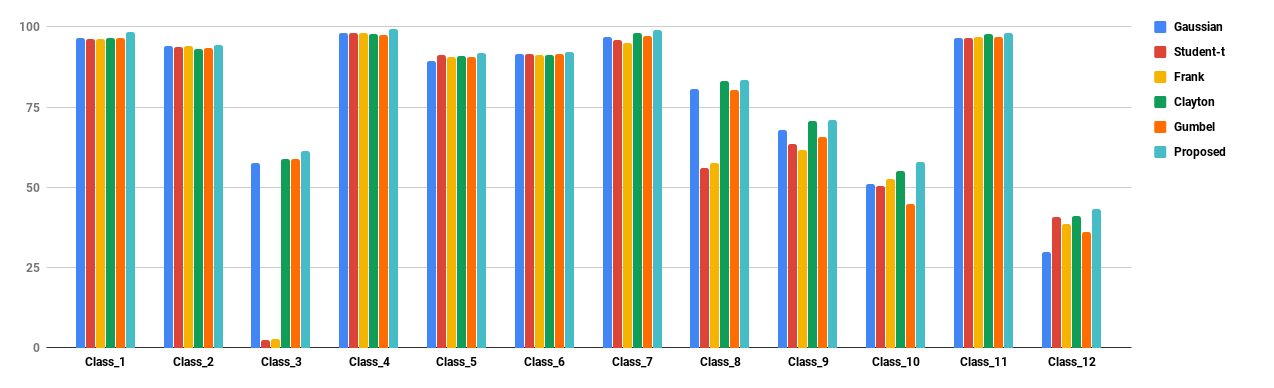} 
		\caption{Comparison of accuracies of single class copula ensembling and our proposed method for each class of CamVid dataset which shows our proposed method has better accuracy for every class.  }
		\label{bar_graph}
	\end{figure*}
	
	\subsection{Performance Measure :} \label{performance}
	In this section, we discuss three performance metrics which are used here, namely, the Overall pixel Accuracy(OA), which is the percentage of total number of correctly classified pixels; Mean class accuracy(CA), which is the mean of all the class accuracies and Mean Intersection over Union (mIOU). the formula for OA is formulated as
	\begin{align*}
	\text{OA} = \frac{\sum_{j=1}^{N}\hat{y}_j==y_j}{N}\times 100 
	\end{align*} 
	For a given class $l$, the class accuracy $CA_{l}$ is defined as \begin{align*}
	\text{CA}_{l} &= \frac{\sum_{j=1}^{N} \hat{y_{j}} == l \land y_{j} == l}{\sum_{j=1}^{N} y_{j} == l}\\
	\text{Hence the mean Class }&\text{Accuray or CA will be } \nonumber \\
	\text{CA} &= \frac{\sum_{l} {CA}_{l} }{l} \times 100\\ 
	\end{align*}For a given class $l$, IOU is defined as 
	\begin{align}
	IOU(l) & = \dfrac{\sum_{j=1}^{N} \hat{y_{j}} == l \land y_{j} == l }{\sum_{j=1}^{N} \hat{y_{j}} == l \lor y_{j} == l }  \label{eq12} \nonumber \\ 
	\text{Hence the mean IOU is } & \nonumber \\
	mIOU & = \frac{\sum_{l} IOU(l)}{l} \nonumber
	\end{align}
	Where $\land$ is 'and', $\lor$ is 'or' operation, $\hat{y_{j}}$ is the prediction and $y_{j}$ is the ground truth of the pixel j. Where $N$ is the total number of pixels in the entire dataset.
	
	\subsection{Other State-of-the-art ensemble methods for comparison} The performance of the proposed ECD3 ensemble model is compared with three other popular classifier ensemble methods namely LOP \cite{lop}, majority voting \cite{42,43,44,45}, and fusion based on simple logit model \cite{logit}. \\ LOP or linear opinion pool is just weighted average of all confidence score obtained from different classifiers i.e. if we have $M$ classifiers and $p_j$ is a confidence score obtained from classifier $j$ then the resultant confidence score $p$ will be $$ p = \sum_{i = 1}^{M} w_ip_i$$ where $\sum_{i = 1}^{M}w_i = 1$. 
	\\
	In majority voting, we determine the optimal decision according to the decision made by each base architectures. If the majority of classifiers decides that a certain pixel belongs to a particular class then that pixel will belong to that particular class. But if each classifier gives different results on the same sample then the ultimate decision will be made by comparing their confidence score of that pixel. \\
	On the other hand, in case of fusion based on a simple logit model, the fusion rule is defined as $$p=\frac{\left[\prod_{i=1}^{M}\left(\frac{p_{i}}{1-p_{i}}\right)^{1/M}\right]^{a}}{1+\left[\prod_{i=1}^{M}\left(\frac{p_{i}}{1-p_{i}}\right)^{1 / M}\right]^{a}}$$
	
	\subsection{Result and Analysis :} \label{result}
	In this section, we have discussed the detailed result and analysis of our experiments. As mentioned in section \ref{experiment}, we have used the output results obtained from SegNet \cite{BadrinarayananSegNet:Segmentation}, PSPNet\cite{ZhaoPyramidNetwork}, Tiramisu \cite{JegouTheSegmentation} as belief scores to ensemble further. Note that we did not use any pre-trained model or transfer learning for those deep learning networks. We have trained those base models from scratch on CamVid and ICCV09 datasets using the PyTorch environment. We have also avoided the use of the data augmentation technique to show the improvement of our proposed ensembling technique with the original dataset. It keeps our experimental protocol simple and robust with minimal data. It is because the motto of the paper is not to create benchmark results on the  CamVid and ICCV09 dataset, rather establish the effectiveness of class-specific Copula for multi-class classification problems for semantic image segmentation purposes, which is a challenging task.
	\begin{table}[h]
		\caption{The best fitted copula functions for each class determined by compairing fitting statistics for CamVid dataset}
		\label{class_wise_copula}
		\begin{tabular}{p{0.3\linewidth}p{0.6\linewidth}}
			\toprule
			\textbf{Best Fitted Copula} & \textbf{Classes} \\ \midrule 
			\textbf{Gaussian} & Sky, Building, \\
			\textbf{Student-t} & Road, Side-walk \\
			\textbf{Frank} & Tree \\
			\textbf{Clayton} & Sign-Symbol, Fence, Car, Pedestrian, void \\
			\textbf{Gumbel} & Column-Pole, Bi-Cyclist \\ \bottomrule 
		\end{tabular}
		
	\end{table}
	\begin{table}[h]
		
		\caption{Results of Ensembling data from SegNet, PSPNet and Tiramisu on CamVid dataset }
		\label{main_table}
		\begin{tabular}{p{0.22\linewidth}|p{0.2\linewidth}p{0.2\linewidth}p{0.2\linewidth}}
			\toprule
			\textbf{Architechtures}     & \textbf{Overall Accuracy} & \textbf{Mean Accuracy} & \textbf{Mean IOU } \\ 
			\midrule 
			SegNet\cite{BadrinarayananSegNet:Segmentation} & 84.700303 & 49.519581 & 0.41825 \\
			PSPNet\cite{ZhaoPyramidNetwork} & 92.818602 & 78.782833 & 0.663291  \\ 
			Tiramisu\cite{JegouTheSegmentation} & 91.637061 & 77.221778 & 0.637337  \\ 
			\midrule
			LOP\cite{lop} & 92.608401 & 81.615978 & 0.635698 \\
			Majority\_voting\cite{42} & 92.8761 & 80.979356 & 0.652887 \\ 
			Logit\cite{logit} & 92.608401 & 81.615978 & 0.635698  \\ 
			\midrule
			Gaussian & 90.913687 & 79.188869 & 0.60419  \\ 
			Student-t & 88.045649 & 73.047487 & 0.556318  \\ 
			Frank & 87.969128 & 72.89565 & 0.552939 \\ 
			Clayton & 91.90342 & 81.650784 & 0.65254\\ 
			Gumbel & 91.119866 & 79.080911 & 0.579787   \\ 
			\midrule
			Proposed & \textbf{93.091532} & \textbf{82.559746} & \textbf{0.672016}  \\
			\bottomrule
		\end{tabular}
	\end{table}
	
	\begin{table*}[h]
		
		\caption{Results of Ensembling data by taking two base model at a time on CamVid dataset}
		\label{big_main_table}
		\begin{tabularx}{\textwidth}{p{0.1\linewidth}|p{0.07\linewidth}p{0.07\linewidth}p{0.07\linewidth}|p{0.07\linewidth}p{0.07\linewidth}p{0.07\linewidth}|p{0.07\linewidth}p{0.07\linewidth}p{0.07\linewidth}}
			\toprule
			\textbf{Combination}& \multicolumn{3}{|c} {\textbf{SegNet\cite{BadrinarayananSegNet:Segmentation}+PSPNet\cite{ZhaoPyramidNetwork}}} &  \multicolumn{3}{|c} {\textbf{SegNet\cite{BadrinarayananSegNet:Segmentation}+Tiramisu\cite{JegouTheSegmentation}}} &   \multicolumn{3}{|c} {\textbf{PSPNet\cite{ZhaoPyramidNetwork}+Tiramisu\cite{JegouTheSegmentation}}} \\
			\midrule 
			& \textbf{Overall Accuracy}  & \textbf{Mean  Accuracy}  &\textbf{ Mean IOU} &\textbf{Overall Accuracy}  & \textbf{Mean  Accuracy}  & \textbf{Mean IOU}&\textbf{Overall Accuracy}  & \textbf{Mean  Accuracy}   & \textbf{Mean IOU}\\ 
			\midrule
			
			LOP\cite{lop} & 90.44951 &\textbf{78.88017 }&0.554523& 90.203623 & 77.322218 & 0.544379& 93.570384 & 82.819487 & 0.674216  \\ 
			MV\cite{42} & 91.759322 &     77.205833 & 0.602118 & 90.790956 & 74.958566 & 0.571028 & 93.34285 & 81.038866 & 0.67964  \\ 
			Logit\cite{logit} & 91.613707 &    78.509435&0.595715 & 90.684692 & 75.856698 & 0.564043    & 93.422855 & 81.80753 & 0.68286 \\ 
			\midrule
			Gaussian & 91.491365 &77.310408 &0.643167 & 90.947187 & 76.684308 & 0.538733 & 92.338565 & 81.417598 & 0.660313 \\ 
			Student-t & 92.179364 &    70.905434&    0.655451 & 88.871516 & 73.510087 & 0.513902 & 90.442771 & 77.845975 & 0.638756\\ 
			Frank & 90.739635 &    77.813193 & 0.564572 & 90.126209 & 76.568786 & 0.533771 & 93.013075 & 82.535629 & 0.657386  \\ 
			Clayton & 91.828268 &78.703963&    0.648968 & 90.060454 & 77.022918 & 0.525882 & 93.712067 & \textbf{83.200367} & 0.653193\\
			Gumbel & 90.928037 &76.683425&    0.551151  & 89.851445 & 76.859128 & 0.520351& 89.76712 & 80.05377 & 0.629215  \\ 
			\midrule
			Proposed & \textbf{93.659201} &77.566497 & \textbf{    0.679266} & \textbf{91.967287} & \textbf{77.364829} & \textbf{0.672016} & \textbf{93.900994} & 81.784975 & \textbf{0.693607}\\ 
			\bottomrule
		\end{tabularx}
	\end{table*}
	\begin{figure}[h!]
		
		\centering\includegraphics[width=1\linewidth]{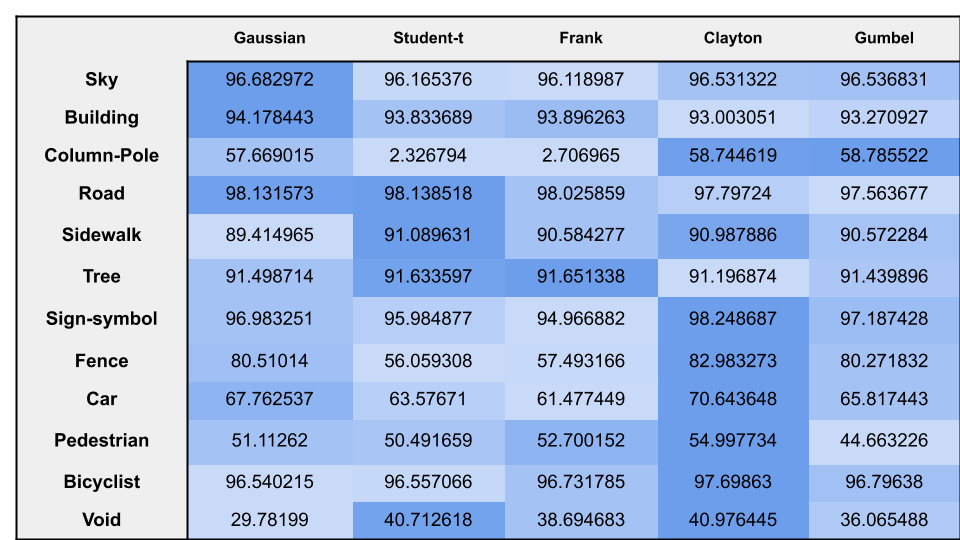} 
		\caption{Performance of single class copula ensembling technique for each class in CamVid dataset.}
		\label{color}
	\end{figure}
	\subsubsection{Results on CamVid Dataset }
	\begin{figure}[h!]
		\centering\includegraphics[width=1\linewidth]{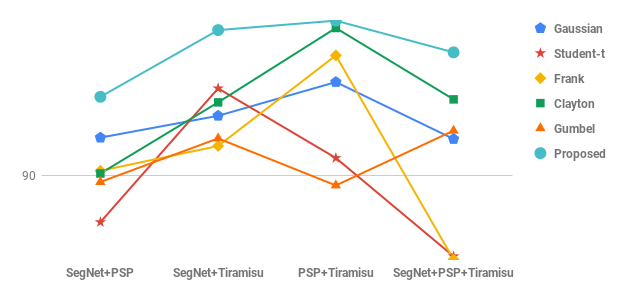} 
		\caption{ Comparision of overall accuracies for each combination between SegNet, PSPNet and Tiramisu net with single class copula ensembling and our proposed method on CamVid dataset.}
		\label{accuracy}
	\end{figure}
	First, we need to get the belief scores of the three deep learning methods after pixel-wise distribution in their associated classes, i.e., $12$ classes for CamVid data set, to obtain the results of the proposed EDC3.   The softmax function is used at the end of the three deep learning techniques to obtain those probabilities.  These probability distributions of the training models are used to improve further using Copula based ensemble method. From the pixel level distribution, the determine the best-fitted Copula function for each class empirically by fitting with the five different Copula functions.
	
	The best fitted copula for each class is selected by comparing the fitting statistics as mentioned in section \ref{Measuring the fit}. The observation on  different fitting statistics of the Copula functions are shown in Table \ref{class_wise_copula}. From the table we can observe that \textbf{Gaussian copula} is the best-fitted copula for classes \textbf{Sky and Building}, the \textbf{Student-t copula} is for the classes \textbf{Road and Sidewalk}; \textbf{Frank copula} for class \textbf{Tree}; \textbf{Gumbel} is for classes C\textbf{olumn-Pole and Bicyclist;} finally \textbf{Clayton copula} is the best fitted one for rest of the classes. 
	After obtaining the class-specific best-fitted copulas, we have ensembled our validation data using \textbf{Algorithm 1} to obtain the fused belief score. We have compared our proposed approach with the three popular state-of-the-art ensembling models on the same dataset, and obtained results of our experiments along with benchmarks are shown in Table \ref{main_table}. We can observe from Table \ref{main_table} that the majority voting technique performs better than each base model in terms of overall accuracy. On the other hand, all the benchmark models and single class copula models perform significantly better in terms of mean accuracy. However, our proposed class-specific copula method outperforms all the above models in terms of overall accuracy, mean accuracy, and mean IOUs. 
	\begin{table}[h]
		\caption{Results of Ensembling data from SegNet, PSPNet and Tiramisu on  ICCV09 dataset )}
		\label{main_iccv09}
		\begin{tabular}{p{0.22\linewidth}|p{0.2\linewidth}p{0.2\linewidth}p{0.2\linewidth}}
			\toprule

			\textbf{Architechtures }     & \textbf{Overall Accuracy} & \textbf{Mean Accuracy} & \textbf{Mean IOU} \\ 
			\midrule
			
			SegNet\cite{BadrinarayananSegNet:Segmentation} &  75.336665 & 58.313291 & 0.476233 \\
			PSPNet\cite{ZhaoPyramidNetwork} &  83.22131 & 66.745377 & 0.525069 \\ 
			Tiramisu\cite{JegouTheSegmentation} &  66.064429 & 55.15124 & 0.361695 \\ 
			\midrule
			LOP\cite{lop} &  82.909654 & \textbf{68.870868 }& 0.538002 \\
			Majority\_voting\cite{42} &  82.632492 & 0.540236 & 0.511325 \\ 
			Logit\cite{logit} &  83.354601 & 68.274124 & 0.53233 \\ 
			\midrule
			Gaussian &  82.862361 & 67.139727 & 0.519109 \\ 
			Student-t &  82.864234 & 67.119442 & 0.519202 \\ 
			Frank &  82.830263 & 67.029158 & 0.518913 \\ 
			Clayton &  83.051362 & 67.12981 & 0.517773 \\ 
			Gumbel &  82.882825 & 67.140376 & 0.519423 \\ 
			\midrule
			Proposed & \textbf{83.821968} & 68.326652 & \textbf{0.548254} \\
			\bottomrule
			
		\end{tabular}
	\end{table}    
	The comparisons of ensembling with single copula functions and our proposed technique for each class on the CamVid dataset are shown in Fig. \ref{bar_graph}. From that figure it is clear that \textit{ensembling with class-specific copula function performs better than ensembling with single copula function.}
	
	\begin{table*}[h!]
		\caption{Results of Ensembling data by taking two architecture at a time on ICCV09 dataset }
		\label{big_iccv09}
		\begin{tabularx}{\textwidth}{p{0.11\linewidth}|p{0.07\linewidth}p{0.07\linewidth}p{0.07\linewidth}|p{0.07\linewidth}p{0.07\linewidth}p{0.07\linewidth}|p{0.07\linewidth}p{0.07\linewidth}p{0.07\linewidth}}
			\toprule
			\textbf{Combination}& \multicolumn{3}{|c} {\textbf{SegNet\cite{BadrinarayananSegNet:Segmentation}+PSPNet\cite{ZhaoPyramidNetwork}}} &  \multicolumn{3}{|c} {\textbf{SegNet\cite{BadrinarayananSegNet:Segmentation}+Tiramisu\cite{JegouTheSegmentation}}} &   \multicolumn{3}{|c} {\textbf{PSPNet\cite{ZhaoPyramidNetwork}+Tiramisu\cite{JegouTheSegmentation}}} \\
			\midrule 
			
			& Overall Accuracy & Mean Accuracy & Mean IOU& Overall Accuracy & Mean Accuracy & Mean IOU & Overall Accuracy & Mean Accuracy & Mean IOU \\   
			\midrule
			
			LOP\cite{lop} &  84.015697 &68.435561 &    \textbf{0.550204} &77.233354 & 62.325241 & 0.49395 & 82.318858 & 67.727353 & 0.511844 \\ 
			Majority\_voting\cite{42} & 83.656696 &\textbf{68.462354}&    0.542648& 77.020235 & 62.224943 & 0.4921& 82.302316 & 68.02633 & 0.511325 \\ 
			Logit\cite{logit} &  83.895423 &    68.163356 &0.548655& 77.153738 & 62.274582 & 0.495565 & 82.823065 & \textbf{68.087257 }& 0.51364 \\
			\midrule 
			Gaussian &     73.515115 &    67.475968 &0.479339& 73.267483 & 62.97528 & 0.474277& 82.640709 & 67.815459 & 0.509369 \\ 
			Student-t & 83.268663 &66.771422&0.530653& 77.313422 & 62.715795 & \textbf{0.497809} & 82.64648 & 67.815459 & 0.509369 \\
			Frank &      83.229601 &    66.748022 &0.531364 & 77.347632 & 62.632267 & 0.497337 & 77.347632 & 62.632267 & 0.497337 \\
			Clayton & 83.23804&    66.776492&0.531374& 68.809506 & 64.051407 & 0.459335 & 73.985753 & 67.861775 & 0.461731 \\ 
			Gumbel &  83.277428 & 66.740492 &     0.53075  & 68.62281 & \textbf{64.085221} & 0.453662& 69.993984 & 67.734381 & 0.445879 \\ 
			\midrule
			
			Proposed & \textbf{84.270355} & 66.750306&    0.550144& \textbf{77.357209} & 62.603572 & 0.497515  & \textbf{83.633214} & 68.02635 & \textbf{0.53625} \\
			\bottomrule
		\end{tabularx}
	\end{table*}
	In Fig. \ref{color}, we have given the class-wise accuracies of ensembling with a single copula. In that figure, we color-coded the result along each row such that boldfaced values indicate the maximum accuracies. From Fig.
	\ref{color}, we can see that Gaussian Copula performs better for classes Sky and Building. Whereas Clayton copula performs better for classes Sidewalk, Sign-symbol, Fence, Car, pedestrian, Bi-cyclist. For class Column-pole Clayton and Gumbel performs closely, but the later has slightly better accuracy. In the case of class Road, gaussian and student-t have similar performance, but student-t has slightly better accuracy. At last for class, Tree has very similar performance with student-t and frank copula, but frank's performance slightly better. In conclusion, Fig. \ref{color} gives a very convincing proof that the chosen copula functions for each class (as shown in Table \ref{class_wise_copula}) are pretty accurate.\\
	\begin{figure}[h!]
		\centering\includegraphics[width=0.8\linewidth]{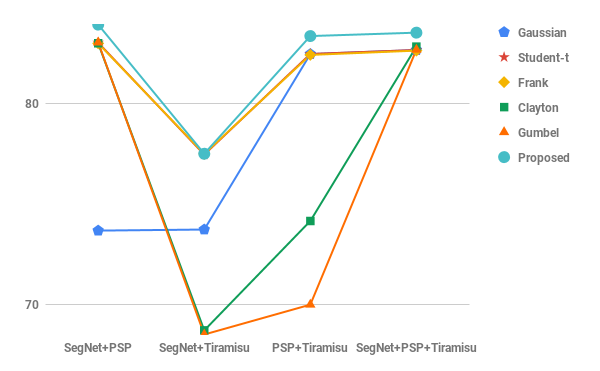} 
		\caption{ Comparision of overall accuracies for each combination between SegNet, PSPNet and Tiramisu net with single class copula ensembling and our proposed method onICCV09 dataset.}
		\label{accuracy_iccv09}
	\end{figure}
	For further analysis, we have shown the numerical results of our experiments performed by taking two classifiers at a time in Table \ref{big_main_table}. The combination of SegNet+PSPNet, our model performs better in terms of overall accuracy and mean IOU, but LOP\cite{lop} has better mean accuracy among all. On the other hand, in SegNet+Tiramisu, our model outperforms the individual base models as well as benchmarks in all performance matrices. In PSPNet+Tiramisu, all the benchmarks, Clayton and frank copula models perform better than the base models, but our proposed models outperform all of them in terms of overall accuracy. Also, note that the overall accuracy of our proposed approach with PSPNet+Tiramisu is better than the combination of SegNet+PSPNet+Tiramisu. That is because the performance of PSPNet and Tiramisu are very similar, and it is expected that the combination of those two will give a better result. On the other hand, SegNet performs poorly compared to the other two. So the combination of SegNet+Tiramisu and SegNet+PSPNet gave significantly lower accuracy. 
	
	\subsubsection{ICCV09 dataset} 
	To validate the robustness of the proposed technique, we have also evaluated our method on another dataset, ICCV09, which consists of 9 different classes. Here "Clayton Copula" is the best-fitted Copula for the majority of the classes, namely Sky, Tree, Grass, Ground, Mountain Object. The other selected Copulas are  Student-t for Building and water, Frank for the unknown class. It is worthy of mentioning that Gaussian Copula, which normally used popularly  \cite{Salinas-GutierrezUsingClassification}  is not performed as the best Copula for this dataset. Even for the dataset CamVid, the Gaussian Copula performs better for only two classes among 12 classes.  From  Table \ref{main_iccv09}, it is observed that a simple Logit model achieves the highest mean accuracy. On the other hand, our proposed model has achieved the highest overall accuracy and mean IOU. The performance is better over the base models as well as their combinations based on single Copula. 
	
	We have also experimented with our techniques with pairwise combinations of three deep learning models. From Table \ref{big_iccv09},  it can be observed that for each Copula except Gaussian, Segnet+PSPNet combination with single Copula function performs better than the other combinations using the same Copula function. The performance of the Segnet+PSPNet is also better for the proposed Copula.  As  PSPNet performs better than the other two networks in their base model, hence the combination with PSPNet gives better performance than the combination with the combination without PSPNet.
	
	\begin{figure*}[h!]
		\captionsetup[subfigure]{slc=off,margin={1cm,0cm}}
		\parbox[b]{0.05\textwidth}{\rotatebox{90}{ \quad Samples}}
		\parbox[b]{0.15\textwidth}{\includegraphics[width=\linewidth]{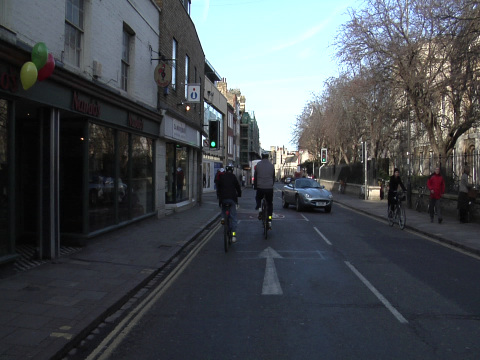}}%
		\parbox[b]{0.15\textwidth}{\includegraphics[width=\linewidth]{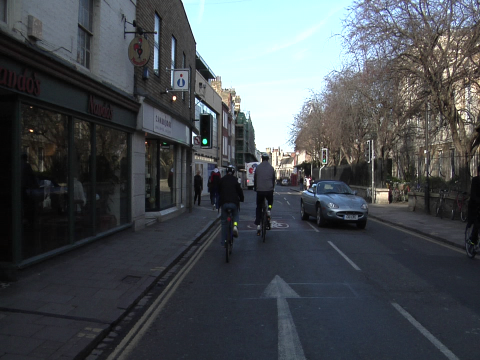}}%
		\parbox[b]{0.15\textwidth}{\includegraphics[width=\linewidth]{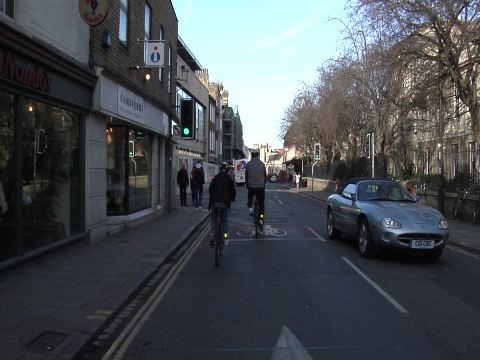}}
		\parbox[b]{0.15\textwidth}{\includegraphics[width=\linewidth]{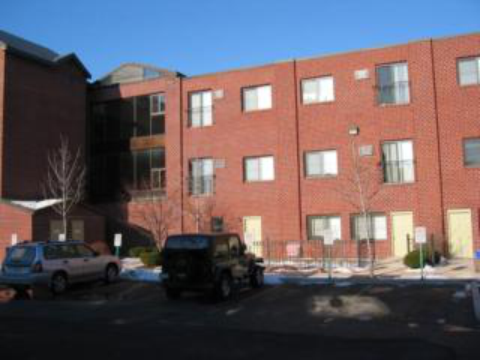}}%
		\parbox[b]{0.15\textwidth}{\includegraphics[width=\linewidth]{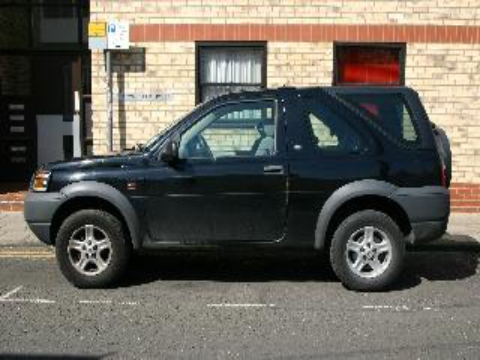}}%
		\parbox[b]{0.15\textwidth}{\includegraphics[width=\linewidth]{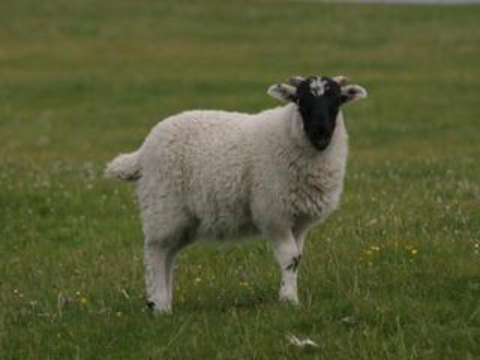}}\\
		\parbox[b]{0.05\textwidth}{\rotatebox{90}{ Annotations}}
		\parbox[b]{0.15\textwidth}{\includegraphics[width=\linewidth]{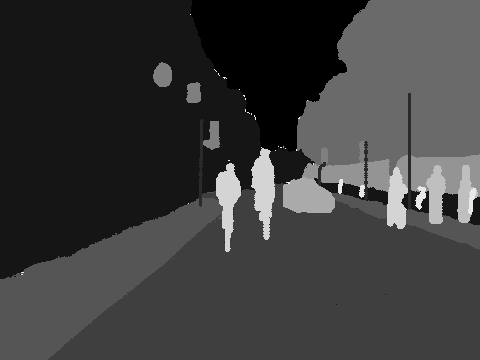}}%
		\parbox[b]{0.15\textwidth}{\includegraphics[width=\linewidth]{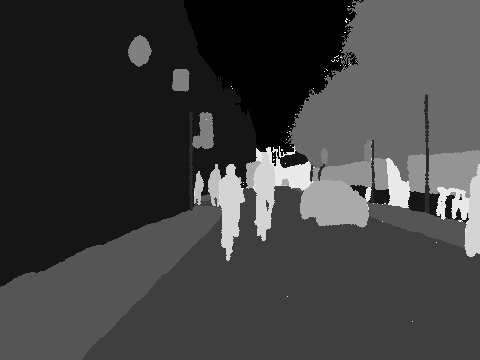}}%
		\parbox[b]{0.15\textwidth}{\includegraphics[width=\linewidth]{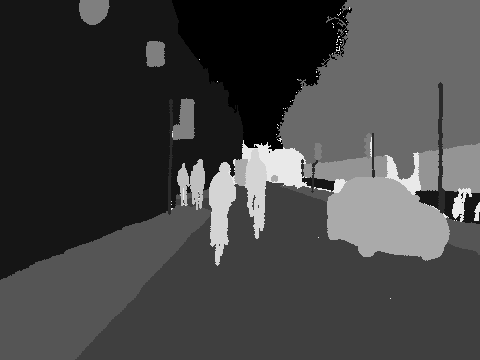}}
		\parbox[b]{0.15\textwidth}{\includegraphics[width=\linewidth]{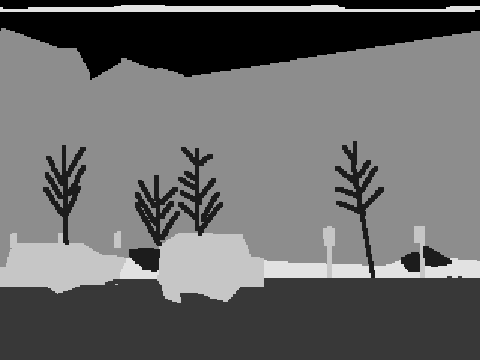}}%
		\parbox[b]{0.15\textwidth}{\includegraphics[width=\linewidth]{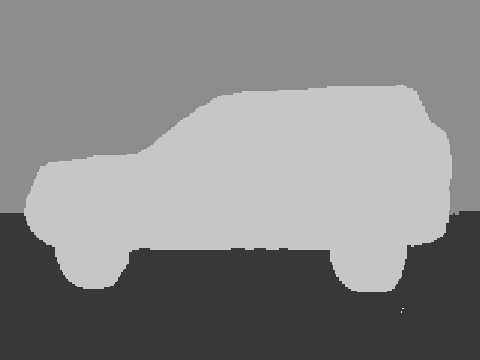}}%
		\parbox[b]{0.15\textwidth}{\includegraphics[width=\linewidth]{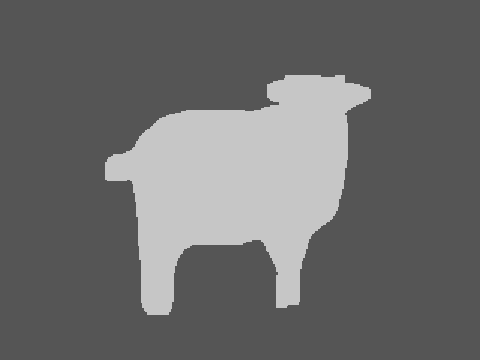}}\\
		\parbox[b]{0.05\textwidth}{\rotatebox{90}{ \quad SegNet}}
		\parbox[b]{0.15\textwidth}{\includegraphics[width=\linewidth]{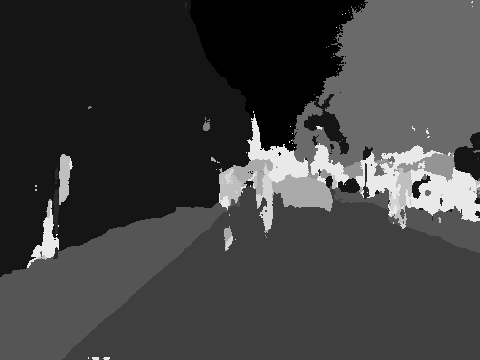}}%
		\parbox[b]{0.15\textwidth}{\includegraphics[width=\linewidth]{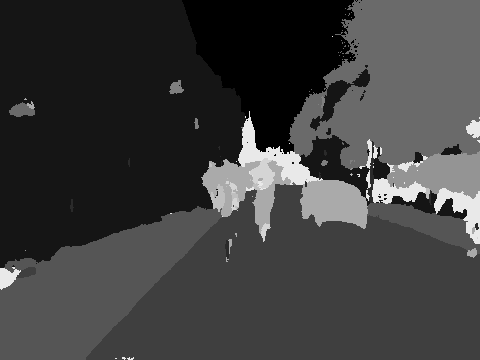}}%
		\parbox[b]{0.15\textwidth}{\includegraphics[width=\linewidth]{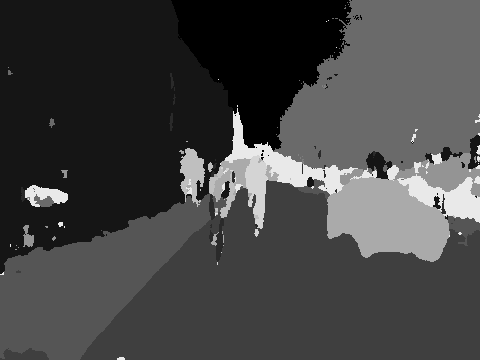}}
		\parbox[b]{0.15\textwidth}{\includegraphics[width=\linewidth]{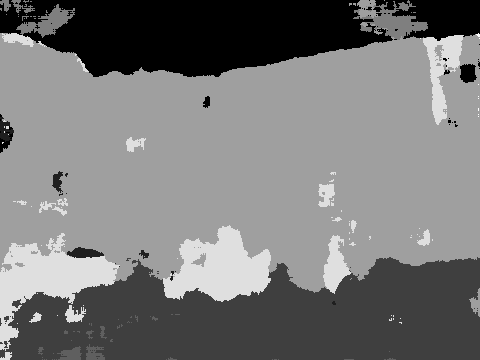}}%
		\parbox[b]{0.15\textwidth}{\includegraphics[width=\linewidth]{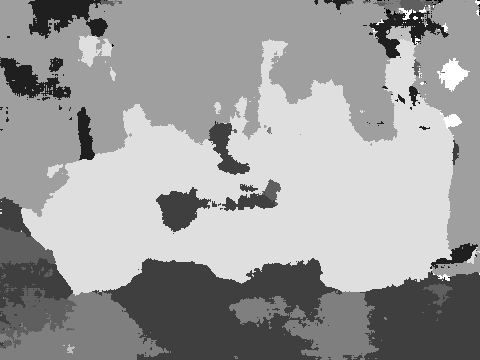}}%
		\parbox[b]{0.15\textwidth}{\includegraphics[width=\linewidth]{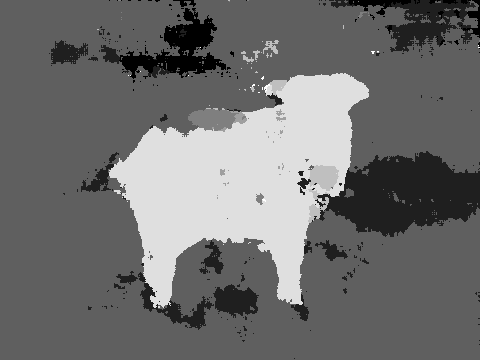}}%
		\\
		\parbox[b]{0.05\textwidth}{\rotatebox{90}{ \quad PSPNet}}
		\parbox[b]{0.15\textwidth}{\includegraphics[width=\linewidth]{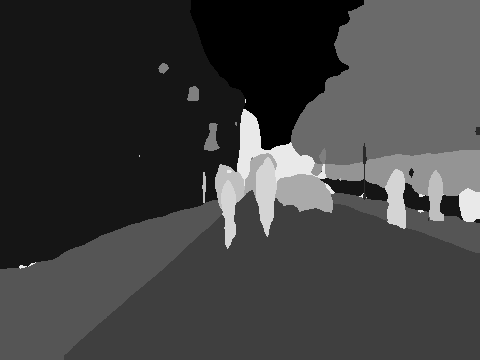}}%
		\parbox[b]{0.15\textwidth}{\includegraphics[width=\linewidth]{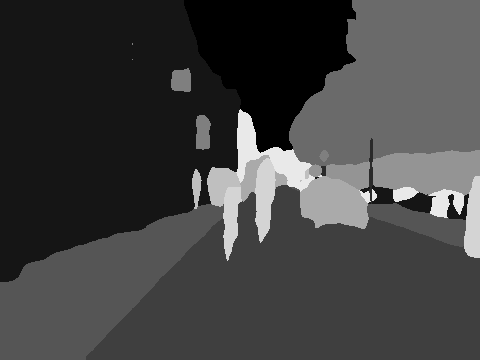}}%
		\parbox[b]{0.15\textwidth}{\includegraphics[width=\linewidth]{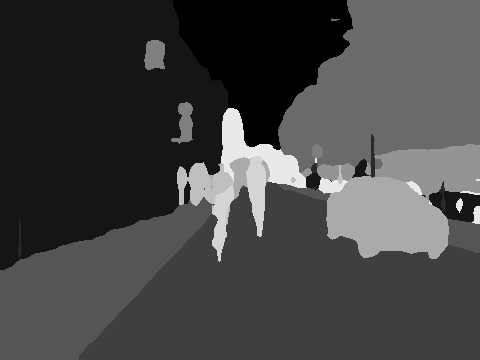}}
		\parbox[b]{0.15\textwidth}{\includegraphics[width=\linewidth]{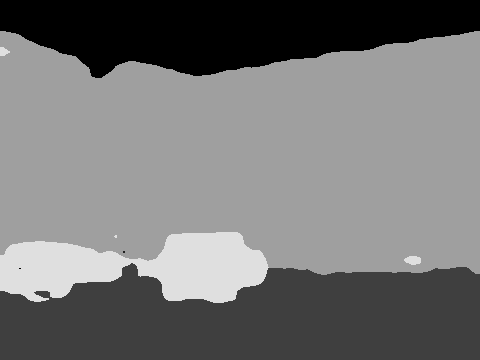}}%
		\parbox[b]{0.15\textwidth}{\includegraphics[width=\linewidth]{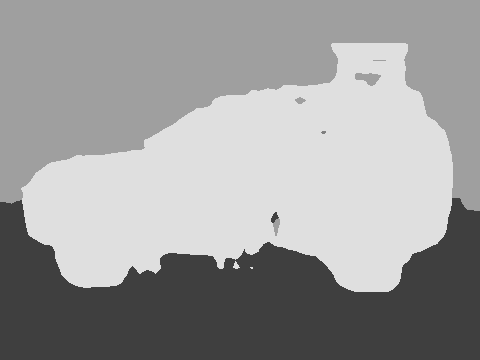}}%
		\parbox[b]{0.15\textwidth}{\includegraphics[width=\linewidth]{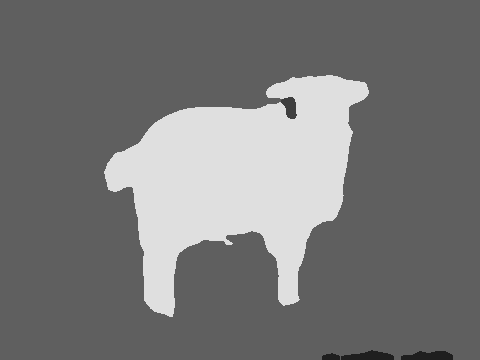}}\\
		\parbox[b]{0.05\textwidth}{\rotatebox{90}{\quad  Tiramisu}}
		\parbox[b]{0.15\textwidth}{\includegraphics[width=\linewidth]{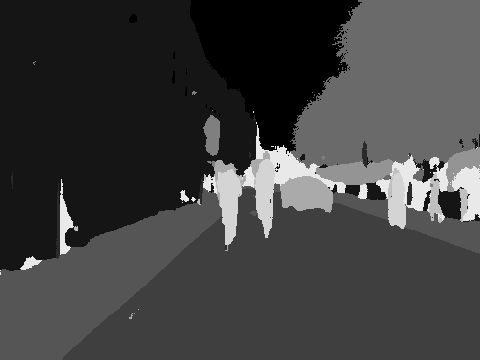}}%
		\parbox[b]{0.15\textwidth}{\includegraphics[width=\linewidth]{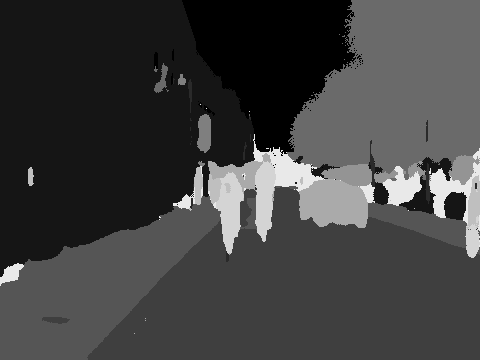}}%
		\parbox[b]{0.15\textwidth}{\includegraphics[width=\linewidth]{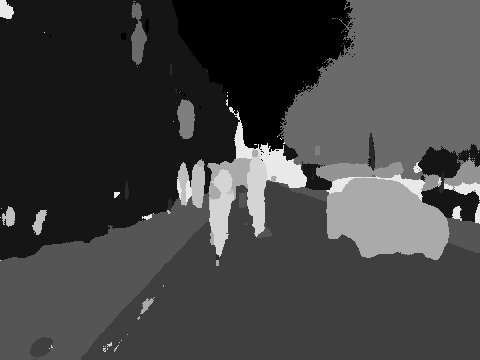}}
		\parbox[b]{0.15\textwidth}{\includegraphics[width=\linewidth]{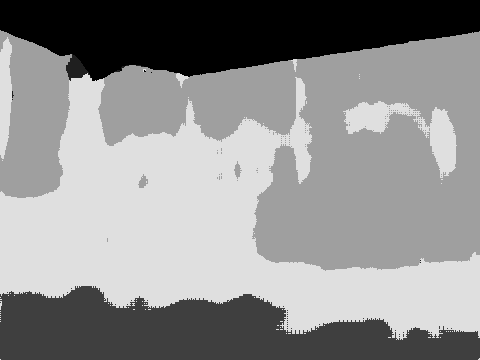}}%
		\parbox[b]{0.15\textwidth}{\includegraphics[width=\linewidth]{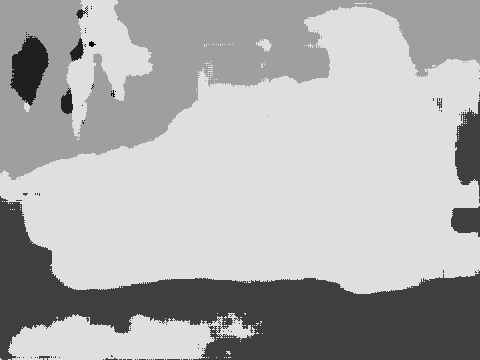}}%
		\parbox[b]{0.15\textwidth}{\includegraphics[width=\linewidth]{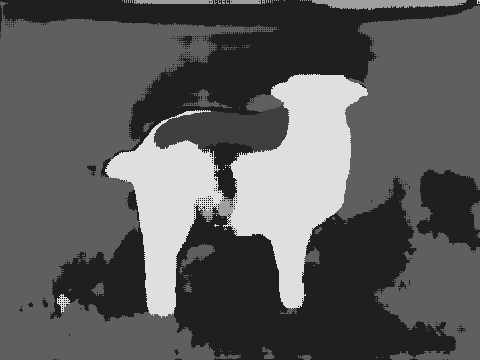}}\\
		\parbox[b]{0.05\textwidth}{\rotatebox{90}{\centering \quad \qquad Proposed}}
		\parbox[b]{0.15\textwidth}{\includegraphics[width=\linewidth]{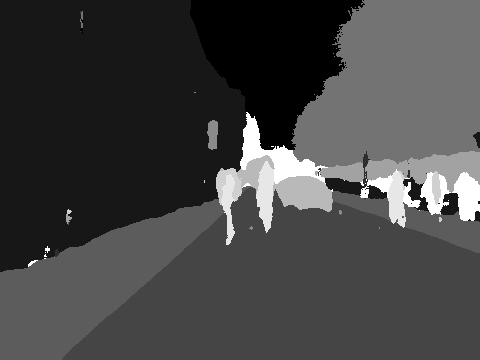} \subcaption{}}%
		\parbox[b]{0.15\textwidth}{\includegraphics[width=\linewidth]{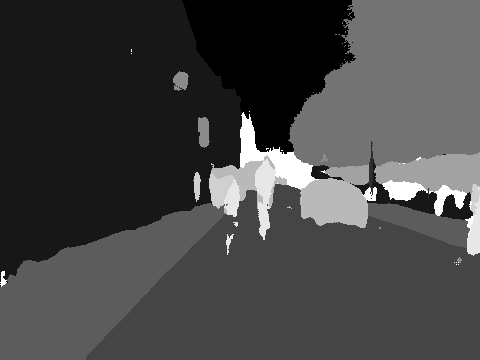}\subcaption{}}%
		\parbox[b]{0.15\textwidth}{\includegraphics[width=\linewidth]{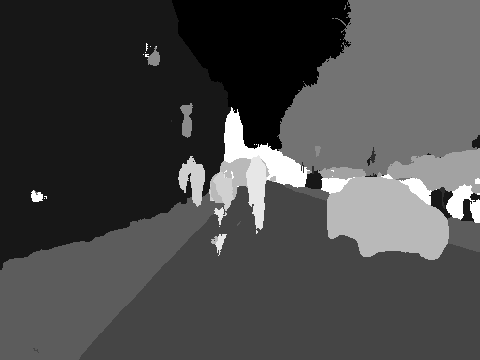}\subcaption{}}
		\parbox[b]{0.15\textwidth}{\includegraphics[width=\linewidth]{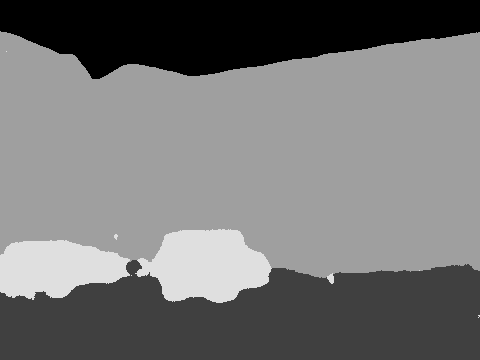}\subcaption{}}%
		\parbox[b]{0.15\textwidth}{\includegraphics[width=\linewidth]{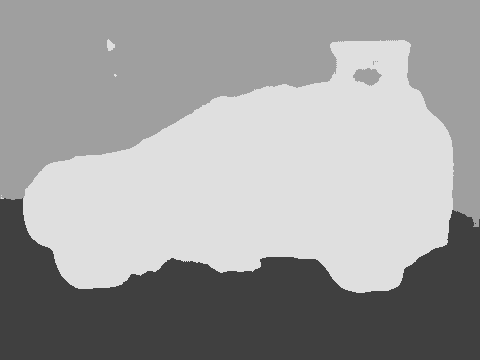}\subcaption{}}%
		\parbox[b]{0.15\textwidth}{\includegraphics[width=\linewidth]{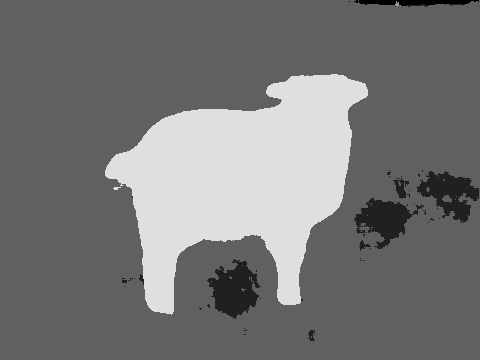}\subcaption{}}
		\caption{Visual representation of performances of our proposed model with the base models on CamVid and ICCV09 datasets. The images indexed with (a),(b),(c) are CamVid samples and (d),(e),(f) are ICCV09 samples}
	\end{figure*}
	
	As shown in Fig. \ref{bar_graph} Our model can perform significantly better than ensembling with single copula functions not only for each class of a dataset but also in terms of overall accuracy, mean IOU, and mean accuracy. 
	The results in table \ref{main_table}, \ref{big_main_table} indicates that our model can perform better if we ensemble classifiers which have similar performances. Also, results in  \ref{main_iccv09} and \ref{big_iccv09} show that mean accuracy and mean IOU can get affected by the ensembling of classifiers having significantly different performance. However, Fig. \ref{accuracy} and Fig. \ref{accuracy_iccv09} indicates that in every combination, our proposed model can boost up the overall pixel accuracy.  Since we are dealing with multi-class classification on the pixel level, the estimation of marginal distribution and determining the best-fitted copula family for each class are time-consuming, which can be addressed in future work.
	
	\section{Conclusion} \label{conclusion}
 We have proposed EDC3: \textbf{E}nsemble of \textbf{D}eep-\textbf{C}lassifer using \textbf{C}lass specific \textbf{C}opula functions to improve semantic image segmentation. Our proposed ensemble model can address inter-source statistical dependencies present among different deep learning-based classifiers for image segmentation. The proposed model explored the effectiveness of class-specific Copula, which provides more flexibility than using a single class model for multiclass problems. Though Copula is well known for binary class image segmentation but its property to use multiclass image segmentation hardly used in literature. So multiclass image segmentation using class specific copula function is one of our major contributions in the present work.  The proposed model performs significantly better than ensembling with single Copula functions not only for each class of the datasets but also in terms of overall accuracy, mean IOU, and mean accuracy for semantic image segmentation. It also outperforms some other popular ensemble methods, namely LOP, Majority voting, Fusion based on Logit in most of the performance indices. Since we are dealing with multiclass classification at a pixel level, the estimation of marginal distribution and determining the best-fitted copula family for each class are time-consuming, which needs to be addressed. The technique suggested can also be used for the multiclass segmentation and classification task effectively where statistical dependencies exist among different classifiers.

	\ifCLASSOPTIONcompsoc
	\section*{Acknowledgments}
	\else
	\section*{Acknowledgment}
	\fi

	This work is partially supported by the project order no. SB/S3/EECE/054/2016, dated 25/11/2016, sponsored by SERB (Government of India) and carried out at the Centre for Microprocessor Application for Training Education and Research, CSE Department, Jadavpur University.
	One of the authors would like to thank The Department of Science and Technology for their INSPIRE Fellowship program (IF170641) for the financial support.

	\ifCLASSOPTIONcaptionsoff
	\newpage
	\fi

	\bibliographystyle{IEEEtran}  
	\bibliography{biblo}  \begin{IEEEbiography}[{\includegraphics[width=1in,height=1.25in,clip,keepaspectratio]{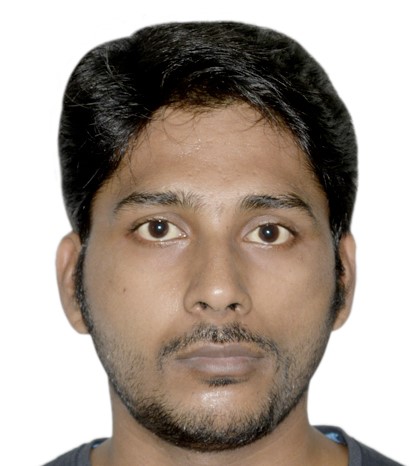}}]{Somenath Kuiry}

\begin{thebibliography}{10}
\providecommand{\url}[1]{#1}
\csname url@samestyle\endcsname
\providecommand{\newblock}{\relax}
\providecommand{\bibinfo}[2]{#2}
\providecommand{\BIBentrySTDinterwordspacing}{\spaceskip=0pt\relax}
\providecommand{\BIBentryALTinterwordstretchfactor}{4}
\providecommand{\BIBentryALTinterwordspacing}{\spaceskip=\fontdimen2\font plus
\BIBentryALTinterwordstretchfactor\fontdimen3\font minus
  \fontdimen4\font\relax}
\providecommand{\BIBforeignlanguage}[2]{{%
\expandafter\ifx\csname l@#1\endcsname\relax
\typeout{** WARNING: IEEEtran.bst: No hyphenation pattern has been}%
\typeout{** loaded for the language `#1'. Using the pattern for}%
\typeout{** the default language instead.}%
\else
\language=\csname l@#1\endcsname
\fi
#2}}
\providecommand{\BIBdecl}{\relax}
\BIBdecl

\bibitem{ghosh2019understanding}
S.~Ghosh, N.~Das, I.~Das, and U.~Maulik, ``Understanding deep learning
  techniques for image segmentation,'' \emph{arXiv preprint arXiv:1907.06119},
  2019.

\bibitem{shi2000normalized}
J.~Shi and J.~Malik, ``Normalized cuts and image segmentation,''
  \emph{Departmental Papers (CIS)}, p. 107, 2000.

\bibitem{achanta2010slic}
R.~Achanta, A.~Shaji, K.~Smith, A.~Lucchi, P.~Fua, and S.~S{\"u}sstrunk, ``Slic
  superpixels,'' Tech. Rep., 2010.

\bibitem{long2015fully}
J.~Long, E.~Shelhamer, and T.~Darrell, ``Fully convolutional networks for
  semantic segmentation,'' in \emph{Proceedings of the IEEE conference on
  computer vision and pattern recognition}, 2015, pp. 3431--3440.

\bibitem{chen2014semantic}
L.-C. Chen, G.~Papandreou, I.~Kokkinos, K.~Murphy, and A.~L. Yuille, ``Semantic
  image segmentation with deep convolutional nets and fully connected crfs,''
  \emph{arXiv preprint arXiv:1412.7062}, 2014.

\bibitem{auto}
{\c{C}}.~Kaymak and A.~U{\c{c}}ar, ``A brief survey and an application of
  semantic image segmentation for autonomous driving,'' in \emph{Handbook of
  Deep Learning Applications}.\hskip 1em plus 0.5em minus 0.4em\relax Springer,
  2019, pp. 161--200.

\bibitem{doi:10.1146/annurev.bioeng.2.1.315}
\BIBentryALTinterwordspacing
D.~L. Pham, C.~Xu, and J.~L. Prince, ``Current methods in medical image
  segmentation,'' \emph{Annual Review of Biomedical Engineering}, vol.~2,
  no.~1, pp. 315--337, 2000, pMID: 11701515. [Online]. Available:
  \url{https://doi.org/10.1146/annurev.bioeng.2.1.315}
\BIBentrySTDinterwordspacing

\bibitem{forouzanfar2010parameter}
M.~Forouzanfar, N.~Forghani, and M.~Teshnehlab, ``Parameter optimization of
  improved fuzzy c-means clustering algorithm for brain mr image
  segmentation,'' \emph{Engineering Applications of Artificial Intelligence},
  vol.~23, no.~2, pp. 160--168, 2010.

\bibitem{delmerico2011building}
J.~A. Delmerico, P.~David, and J.~J. Corso, ``Building facade detection,
  segmentation, and parameter estimation for mobile robot localization and
  guidance,'' in \emph{2011 IEEE/RSJ International Conference on Intelligent
  Robots and Systems}.\hskip 1em plus 0.5em minus 0.4em\relax IEEE, 2011, pp.
  1632--1639.

\bibitem{liu2018joint}
Z.~Liu, L.~Wang, G.~Hua, Q.~Zhang, Z.~Niu, Y.~Wu, and N.~Zheng, ``Joint video
  object discovery and segmentation by coupled dynamic markov networks,''
  \emph{IEEE Transactions on Image Processing}, vol.~27, no.~12, pp.
  5840--5853, 2018.

\bibitem{wang2018segment}
L.~Wang, X.~Duan, Q.~Zhang, Z.~Niu, G.~Hua, and N.~Zheng, ``Segment-tube:
  Spatio-temporal action localization in untrimmed videos with per-frame
  segmentation,'' \emph{Sensors}, vol.~18, no.~5, p. 1657, 2018.

\bibitem{41}
W.~Nayer, ``Feature based architecture for decision fusion,'' Ph.D.
  dissertation, phD thesis, 2003.

\bibitem{42}
R.~Battiti and A.~M. Colla, ``Democracy in neural nets: Voting schemes for
  classification,'' \emph{Neural Networks}, vol.~7, no.~4, pp. 691--707, 1994.

\bibitem{43}
C.~Ji and S.~Ma, ``Combinations of weak classifiers,'' in \emph{Advances in
  Neural Information Processing Systems}, 1997, pp. 494--500.

\bibitem{44}
L.~Lam and C.~Y. Suen, ``Optimal combinations of pattern classifiers,''
  \emph{Pattern Recognition Letters}, vol.~16, no.~9, pp. 945--954, 1995.

\bibitem{45}
L.~Xu, A.~Krzyzak, and C.~Y. Suen, ``Methods of combining multiple classifiers
  and their applications to handwriting recognition,'' \emph{IEEE transactions
  on systems, man, and cybernetics}, vol.~22, no.~3, pp. 418--435, 1992.

\bibitem{46}
L.~I. Kuncheva, ``A theoretical study on six classifier fusion strategies,''
  \emph{IEEE Transactions on pattern analysis and machine intelligence},
  vol.~24, no.~2, pp. 281--286, 2002.

\bibitem{47}
P.~W. Munro and B.~Parmanto, ``Competition among networks improves committee
  performance,'' in \emph{Advances in Neural Information Processing Systems},
  1997, pp. 592--598.

\bibitem{48}
D.~M. Tax, M.~Van~Breukelen, R.~P. Duin, and J.~Kittler, ``Combining multiple
  classifiers by averaging or by multiplying?'' \emph{Pattern recognition},
  vol.~33, no.~9, pp. 1475--1485, 2000.

\bibitem{49}
T.~K. Ho, J.~J. Hull, and S.~N. Srihari, ``Decision combination in multiple
  classifier systems,'' \emph{IEEE Transactions on Pattern Analysis \& Machine
  Intelligence}, no.~1, pp. 66--75, 1994.

\bibitem{50}
S.-B. Cho and J.~H. Kim, ``Combining multiple neural networks by fuzzy integral
  for robust classification,'' \emph{IEEE Transactions on Systems, Man, and
  Cybernetics}, vol.~25, no.~2, pp. 380--384, 1995.

\bibitem{51}
L.~I. Kuncheva, ``An application of owa operators to the aggregation of
  multiple classification decisions,'' in \emph{The ordered weighted averaging
  operators}.\hskip 1em plus 0.5em minus 0.4em\relax Springer, 1997, pp.
  330--343.

\bibitem{52}
L.~Kuncheva, J.~C. Bezdek, and M.~A. Sutton, ``On combining multiple
  classifiers by fuzzy templates,'' in \emph{1998 Conference of the North
  American Fuzzy Information Processing Society-NAFIPS (Cat. No.
  98TH8353)}.\hskip 1em plus 0.5em minus 0.4em\relax IEEE, 1998, pp. 193--197.

\bibitem{lop}
C.~Genest and K.~J. McConway, ``Allocating the weights in the linear opinion
  pool,'' \emph{Journal of Forecasting}, vol.~9, no.~1, pp. 53--73, 1990.

\bibitem{blp}
R.~Ranjan and T.~Gneiting, ``Combining probability forecasts,'' \emph{Journal
  of the Royal Statistical Society: Series B (Statistical Methodology)},
  vol.~72, no.~1, pp. 71--91, 2010.

\bibitem{logit}
V.~A. Satop{\"a}{\"a}, J.~Baron, D.~P. Foster, B.~A. Mellers, P.~E. Tetlock,
  and L.~H. Ungar, ``Combining multiple probability predictions using a simple
  logit model,'' \emph{International Journal of Forecasting}, vol.~30, no.~2,
  pp. 344--356, 2014.

\bibitem{Ozdemir2017CopulaDependence}
O.~Ozdemir, T.~Allen, S.~Choi, T.~Wimalajeewa, and P.~Varshney, ``{Copula Based
  Classifier Fusion Under Statistical Dependence},'' 2017.

\bibitem{laux2009modelling}
P.~Laux, S.~Wagner, A.~Wagner, J.~Jacobeit, A.~Bardossy, and H.~Kunstmann,
  ``Modelling daily precipitation features in the volta basin of west africa,''
  \emph{International Journal of Climatology: A Journal of the Royal
  Meteorological Society}, vol.~29, no.~7, pp. 937--954, 2009.

\bibitem{Eban2013DynamicSeries}
E.~Eban, G.~Rothschild, A.~Mizrahi, I.~Nelken, and G.~Elidan, ``{Dynamic Copula
  Networks for Modeling Real-valued Time Series},'' Tech. Rep., 2013.

\bibitem{10.1371/journal.pcbi.1000577}
\BIBentryALTinterwordspacing
A.~Onken, S.~Gr{\"{u}}new{\"{a}}lder, M.~H.~J. Munk, and K.~Obermayer,
  ``{Analyzing Short-Term Noise Dependencies of Spike-Counts in Macaque
  Prefrontal Cortex Using Copulas and the Flashlight Transformation},''
  \emph{PLOS Computational Biology}, vol.~5, no.~11, pp. 1--13, 2009. [Online].
  Available: \url{https://doi.org/10.1371/journal.pcbi.1000577}
\BIBentrySTDinterwordspacing

\bibitem{PollanenCurrentDiagnostics}
I.~P{\"{o}}ll{\"{a}}nen, B.~Braithwaite, K.~Haataja, T.~Ikonen, and
  P.~Toivanen, \emph{{Current Analysis Approaches and Performance Needs for
  Whole Slide Image Processing in Breast Cancer Diagnostics}}.

\bibitem{Kao2009MotivatingClimate}
S.~C. Kao, A.~R. Ganguly, and K.~Steinhaeuser, ``{Motivating complex dependence
  structures in data mining: A case study with anomaly detection in climate},''
  in \emph{ICDM Workshops 2009 - IEEE International Conference on Data Mining},
  2009.

\bibitem{wu2014construction}
S.~Wu, ``Construction of asymmetric copulas and its application in
  two-dimensional reliability modelling,'' \emph{European Journal of
  Operational Research}, vol. 238, no.~2, pp. 476--485, 2014.

\bibitem{CuvelierClaytonDecomposition}
E.~Cuvelier and M.~Noirhomme-Fraiture, ``{Clayton copula and mixture
  decomposition},'' Tech. Rep.

\bibitem{Diday2005MixtureFramework}
E.~Diday and M.~Vrac, ``{Mixture decomposition of distributions by copulas in
  the symbolic data analysis framework},'' \emph{Discrete Applied Mathematics},
  2005.

\bibitem{Salinas-GutierrezUsingClassification}
R.~Salinas-Guti{\'{e}}rrez, A.~Hern{\'{a}}ndez-Aguirre, M.~J.~J. Rivera-Meraz,
  and E.~R. Villa-Diharce, ``{Using Gaussian Copulas in Supervised
  Probabilistic Classification},'' Tech. Rep.

\bibitem{Conant-Pablos2009PipeliningTimetabling}
S.~E. Conant-Pablos, D.~J. Maga{\~{n}}a-Lozano, and H.~Terashima-Mar{\'{i}}n,
  ``{Pipelining memetic algorithms, constraint satisfaction, and local search
  for course timetabling},'' in \emph{Lecture Notes in Computer Science
  (including subseries Lecture Notes in Artificial Intelligence and Lecture
  Notes in Bioinformatics)}, 2009.

\bibitem{gtot}
E.~Kole, K.~Koedijk, and M.~Verbeek, ``Testing copulas to model financial
  dependence,'' \emph{Department of Financial Management, RSM Erasmus
  University, Rotterdam, The Netherlands}, 2005.

\bibitem{differentcopula}
R.~B. Nelsen, \emph{An introduction to copulas}.\hskip 1em plus 0.5em minus
  0.4em\relax Springer Science \& Business Media, 2007.

\bibitem{hac}
C.~Savu and M.~Trede, ``Hierarchies of archimedean copulas,''
  \emph{Quantitative Finance}, vol.~10, no.~3, pp. 295--304, 2010.

\bibitem{hac2}
O.~Okhrin and A.~Ristig, ``Hierarchical archimedean copulae: the hac package,''
  SFB 649 Discussion Paper, Tech. Rep., 2012.

\bibitem{ml}
G.~Wei{\ss}, ``Copula parameter estimation by maximum-likelihood and
  minimum-distance estimators: a simulation study,'' \emph{Computational
  Statistics}, vol.~26, no.~1, pp. 31--54, 2011.

\bibitem{ml2}
N.~Gregor, ``Copula parameter estimation by maximum-likelihood and
  minimum-distance estimators,'' \emph{A simulation study}, 2009.

\bibitem{ifm}
H.~Joe and J.~J. Xu, ``The estimation method of inference functions for margins
  for multivariate models,'' 1996.

\bibitem{ifm2}
P.~H. Ferreira and F.~Louzada, ``A modified version of the inference function
  for margins and interval estimation for the bivariate clayton copula sur
  tobit model: An simulation approach,'' \emph{arXiv preprint arXiv:1404.3287},
  2014.

\bibitem{Burnham2004MultimodelSelection}
K.~P. Burnham and D.~R. Anderson, ``{Multimodel inference: Understanding AIC
  and BIC in model selection},'' 2004.

\bibitem{Aho2014ModelBIC}
K.~Aho, D.~Derryberry, and T.~Peterson, ``{Model selection for ecologists: the
  worldviews of AIC and BIC},'' Tech. Rep.~3, 2014.

\bibitem{camvid}
G.~J. Brostow, J.~Fauqueur, and R.~Cipolla, ``Semantic object classes in video:
  A high-definition ground truth database,'' \emph{Pattern Recognition
  Letters}, vol.~30, no.~2, pp. 88--97, 2009.

\bibitem{iccv09}
S.~Gould, R.~Fulton, and D.~Koller, ``Decomposing a scene into geometric and
  semantically consistent regions,'' in \emph{2009 IEEE 12th international
  conference on computer vision}.\hskip 1em plus 0.5em minus 0.4em\relax IEEE,
  2009, pp. 1--8.

\bibitem{BadrinarayananSegNet:Segmentation}
\BIBentryALTinterwordspacing
V.~Badrinarayanan, A.~Kendall, and R.~Cipolla, ``{SegNet: A Deep Convolutional
  Encoder-Decoder Architecture for Image Segmentation},'' Tech. Rep. [Online].
  Available: \url{http://mi.eng.cam.ac.uk/projects/segnet/.}
\BIBentrySTDinterwordspacing

\bibitem{ZhaoPyramidNetwork}
\BIBentryALTinterwordspacing
H.~Zhao, J.~Shi, X.~Qi, X.~Wang, and J.~Jia, ``{Pyramid Scene Parsing
  Network},'' Tech. Rep. [Online]. Available:
  \url{https://github.com/hszhao/PSPNet}
\BIBentrySTDinterwordspacing

\bibitem{JegouTheSegmentation}
\BIBentryALTinterwordspacing
S.~J{\'{e}}gou, M.~Drozdzal, D.~Vazquez, A.~Romero, and Y.~Bengio, ``{The One
  Hundred Layers Tiramisu: Fully Convolutional DenseNets for Semantic
  Segmentation},'' Tech. Rep. [Online]. Available:
  \url{https://github.com/SimJeg/FC-DenseNet}
\BIBentrySTDinterwordspacing

\end{thebibliography}
	received his B.Sc degree in Mathematics from Jagannath Kishore College under Sidho Kanho Birsha University, in 2014. He graduated from the Indian Institute of Technology, Guwahati, in 2016 with an M.Sc. in Mathematics and Computing.  He is currently pursuing his Ph.D. from Jadavpur University since 2017 under the guidance of professor Alaka Das and Dr. Nibaran Das.
	\end{IEEEbiography}
\begin{IEEEbiography}[{\includegraphics[width=1in,height=1.25in,clip,keepaspectratio]{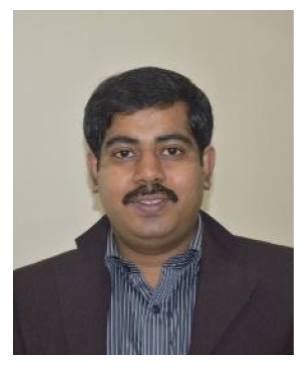}}]{Nibaran Das}
	 received his B.Tech degree in Computer Science and Technology from Kalyani Govt. Engineering College, under Kalyani University, in 2003. He received his M.C.S.E  and Ph. D.(Engg.)  degree from Jadavpur University in 2005 and 2012, respectively. He joined Jadavpur University as a  faculty member in 2006.  His areas of current research interest are Deep Learning, OCR of handwritten text, optimization techniques, and computer vision.  He has been an editor of Bengali monthly magazine “Computer Jagat” since 2005. 
	
\end{IEEEbiography}
\begin{IEEEbiography}[{\includegraphics[width=1in,height=1.25in,clip,keepaspectratio]{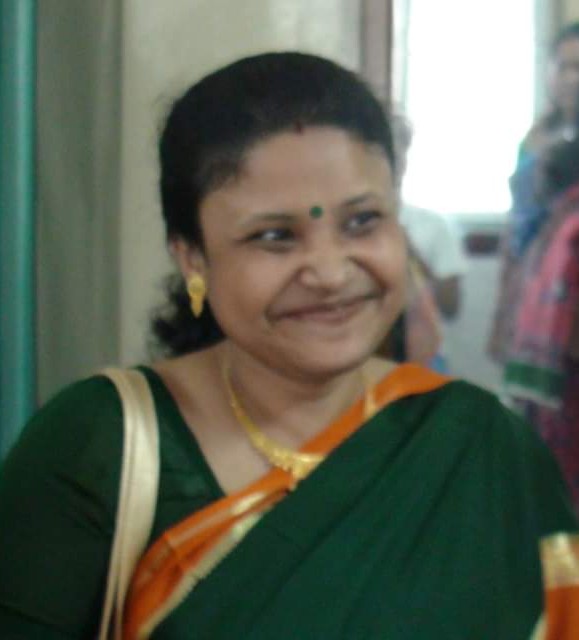}}]{Alaka Das}
received her B.Sc and M.Sc degree in Mathematics from Calcutta University in 1994 and 1996, respectively. She received her Ph. D.(Science)  degree from Jadavpur University in 2006 and also she has been a post-doctorate fellow at the Indian Institute of Technology, Madras, in 2011-2012. She started her teaching career as an assistant professor at Hoogly Women's College in 2001; later, she joined Jadavpur University as a faculty member in 2005.  Her areas of current research interest Nonlinear dynamics, Hydrodynamic and hydromagnetic instability, Dynamical system.  Recently she has been a Course Coordinator of partial differential equations in the Swayam MOOC course of UGC. 	
\end{IEEEbiography}
\begin{IEEEbiography}[{\includegraphics[width=1in,height=1.25in,clip,keepaspectratio]{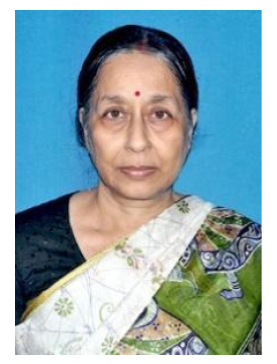}}]{Mita Nasipuri}
received her B.E.Tel.E., M.E.Tel.E., and Ph.D. (Engg.) degrees from Jadavpur University, in 1979, 1981 and 1990, respectively. Prof. Nasipuri has been a faculty member of J.U since 1987. Her current research interest includes image processing, pattern recognition, and multimedia systems. She is a senior member of the IEEE, U.S.A., Fellow of I.E (India) and W.B.A.S.T, Kolkata, India.

\end{IEEEbiography}

\end{document}